\documentclass{article}

    \PassOptionsToPackage{numbers, compress}{natbib}
\usepackage[preprint]{Styles/neurips_2024}

\usepackage[utf8]{inputenc} % allow utf-8 input
\usepackage[T1]{fontenc}    % use 8-bit T1 fonts
\usepackage{hyperref}       % hyperlinks
\usepackage{url}            % simple URL typesetting
\usepackage{booktabs}       % professional-quality tables
\usepackage{amsfonts}       % blackboard math symbols
\usepackage{nicefrac}       % compact symbols for 1/2, etc.
\usepackage{microtype}      % microtypography
\usepackage{xcolor}         % colors

\usepackage{array,multirow,graphicx}
\usepackage{float}
\usepackage{amsmath}
\usepackage{mathtools}
\usepackage{bbm}
\usepackage{hhline}
\usepackage{boldline}
\usepackage{tabularx}
\usepackage{float}
\usepackage{comment}
\usepackage{color}
\usepackage{amssymb}
\usepackage{booktabs}
\usepackage{multicol}
\usepackage{microtype}      % microtypography
\usepackage{xcolor}         % colors
\usepackage{algorithm}
\usepackage{algpseudocode}
\usepackage{hyperref}
\DeclareUrlCommand\url{\color{magenta}}
\hypersetup{
    colorlinks=true,
    linkcolor=blue,
    filecolor=magenta,      
    urlcolor=magenta,
}
\usepackage{url} 
\usepackage{pifont}% http://ctan.org/pkg/pifont
\usepackage{soul} %\st
\newcommand{\xmark}{\ding{55}}%
\usepackage{subcaption}
\usepackage{wrapfig}
\usepackage{lipsum}
\usepackage{booktabs}
\usepackage{xcolor,colortbl}
\usepackage{nicematrix}
\usepackage{kotex}
\usepackage{blindtext}

\newcommand{\Skip}[1]{}

\title{Correlation-Guided Query-Dependency Calibration \\for Video Temporal Grounding} 

\author{%
  WonJun Moon, Sangeek Hyun, SuBeen Lee, Jae-Pil Heo\thanks{Corresponding Author} \\
  Sungkyunkwan University \\
  \texttt{\{wjun0830, hsi1032, leesb7426, jaepilheo\}@g.skku.edu}  \\
}

\begin{document}

\maketitle

\vspace{-0.2cm}
\begin{abstract}
% With our Correlation-Guided DEtection TRansformer~(CG-DETR), we explore the appropriate clip-wise degree of cross-modal interactions and how to exploit such degrees for prediction.
Temporal Grounding is to identify specific moments or highlights from a video corresponding to textual descriptions.
Typical approaches in temporal grounding treat all video clips equally during the encoding process regardless of their semantic relevance with the text query.
Therefore, we propose Correlation-Guided DEtection TRansformer~(CG-DETR), exploring to provide clues for query-associated video clips within the cross-modal attention.
First, we design an adaptive cross-attention with dummy tokens. 
Dummy tokens conditioned by text query take portions of the attention weights, preventing irrelevant video clips from being represented by the text query.
Yet, not all words equally inherit the text query's correlation to video clips. 
Thus, we further guide the cross-attention map by inferring the fine-grained correlation between video clips and words. 
We enable this by learning a joint embedding space for high-level concepts, \textit{i.e}., moment and sentence level, and inferring the clip-word correlation.
Lastly, we exploit the moment-specific characteristics and combine them with the context of each video to form a moment-adaptive saliency detector.
By exploiting the degrees of text engagement in each video clip, it precisely measures the highlightness of each clip.
CG-DETR achieves state-of-the-art results on various benchmarks for temporal grounding.
Codes are available at \href{https://github.com/wjun0830/CGDETR}{github.com/wjun0830/CGDETR}.
% Codes are available at \href{github.com/wjun0830/CGDETR}{github.com/wjun0830/CGDETR}.
% Recent endeavors in video temporal grounding enforce strong cross-modal interactions through attention mechanisms to overcome the modality gap between video and text query.
% However, previous works treat all video clips equally regardless of their semantic relevance with the text query in attention modules.
% In this paper, our goal is to provide clues for query-associated video clips within the cross-modal encoding process.
% With our Correlation-Guided DEtection TRansformer~(CG-DETR), we explore the appropriate clip-wise degree of cross-modal interactions and how to exploit such degrees for prediction.
% First, we design an adaptive cross-attention layer with dummy tokens. 
% Dummy tokens conditioned by text query take a portion of the attention weights, preventing irrelevant video clips from being represented by the text query.
% Yet, not all word tokens equally inherit the text query's correlation to video clips. 
% Thus, we further guide the cross-attention map by inferring the fine-grained correlation between video clips and words. 
% We enable this by learning a joint embedding space for high-level concepts, \textit{i.e}., moment and sentence level, and inferring the clip-word correlation.
% Lastly, we use a moment-adaptive saliency detector to exploit each video clip's degrees of text engagement.
% We validate the superiority of CG-DETR with the state-of-the-art results on various benchmarks for both moment retrieval and highlight detection.
\end{abstract}
\vspace{-0.2cm}
% \blfootnote{
% $^\star$ Corresponding author
% }
% \vspace{-0.1cm}

\vspace{-0.2cm}
\section{Introduction}
\vspace{-0.1cm}
%%%%%%%%%%%% Intro
% VMR / HD는 어떤 task이다.
% Video data is becoming more popular for various purposes, e.g., instructional, entertainment, and daily life recording, due to its semantic richness.On the other hand, as its property of semantic richness also brings challenges in browsing the details, we observe the skyrocketing demand for short clips and highlights. In light of this, the need for video grounding is reaching its peak.

\begin{figure}[t!]
    \centering
    \vspace{-0.6cm}
    \includegraphics[width=0.84\columnwidth]{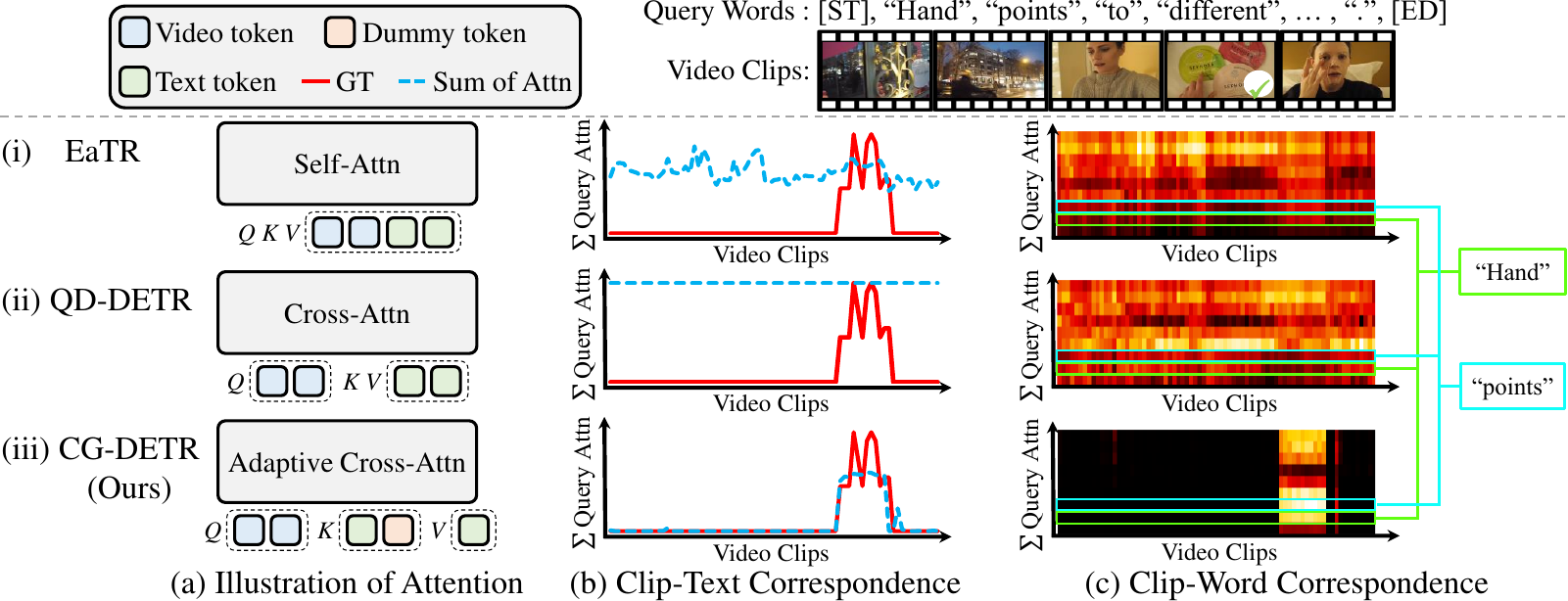}
    \vspace{-0.2cm}
    \caption{
    Comparison of degrees of text-to-video correlation in attention layers.
    In the middle column~(b), we compare the clip-wise correspondence score to the text query~(sum of attention weights over all words) with its corresponding GT~(saliency scores).
    % In the middle row, we compare ground-truth~(GT) saliency scores with the clip-wise correspondence to the text query~(sum of attention weights to every word).
    While the use of (i) self-attention or (ii) cross-attention fails to distinguish target clips based on the degree of cross-modal attention, (iii) ours with adaptive cross-attention exhibits a high activation level for the text query to attend only the query-relevant clips since the dummies occupy a portion of the attention degree on irrelevant clips. 
    We also investigate the fine-grained correlation between clips and words in column~(c). 
    Despite the absence of word-level supervision, ours learns to attend more to salient words.
    % Observation on the degree of text-to-video attention between methods by forwarding video with a paired query.
    % Whereas employing (b)~self-attention and (c)~cross-attention layers cannot distinguish the target clips with the degree of multimodal attention, the text query in ours~(a) is highly activated to attend video clips only when the clips are query-relevant.
    % At the bottom, we also explore the fine-grained correlation in the attention layer between clips and words.
    % Under unavailable supervision, ours learn to identify the salient words.
    }
    \vspace{-0.6cm}
    \label{fig:motiv_figure}
\end{figure}

Video temporal grounding is to detect temporal moments that align with user-specified language requests.
Aside from the moment localization, recent works also explore how well each video clip corresponds with the text query.
To address these tasks, it is crucial to align the representation space across modalities.
In this regard, the use of transformers~\cite{vaswani2017attention} has become a common approach that facilitates easy integration of multimodal representations. 
While some studies~\cite{univtg, umt} use transformer encoders to establish a shared embedding space, others also leverage detection transformers for making predictions~\cite{momentdetr, xu2023mh}.
% Recently, QD-DETR~\cite{qddetr} highlighted previous works' insufficient text reflection in predictions and suggested explicitly forcing text engagement with cross-attention.
Recently, QD-DETR~\cite{qddetr} pointed out the inadequacy of text reflection in predictions made by previous models and suggested explicitly forcing text engagement with cross-attention.
Yet, in moment retrieval scenarios where users seek specific video segments, we claim that an undifferentiated degree of text-to-video attention~(text involvement in video clips), as shown in Fig.~\ref{fig:motiv_figure}~(b), also significantly contributes to inadequate text reflection.
% Yet, in moment retrieval scenarios where users seek specific video segments, we claim that undifferentiated text involvement across all video clips also significantly contributes to inadequate text reflection.
Specifically, in Fig.~\ref{fig:motiv_figure}, we display the clip-wise degree of cross-modal attention where we observe the text engagement in video encodings within (i)~self-attention and (ii)~cross-attention is not differentiated between text-relevant clips and text-irrelevant clips.
This phenomenon is observed in terms of not only between the video and the entire text~(b) but also between the video and each word in the query~(c).
Thus, the model is not given any hint of the relevance between the text query and video clips.
For the importance of the adaptive degree in attention, we examine the positive relationship between the alignment of clip-wise query attention degrees with the corresponding ground truth~(GT) saliency scores and their impact on performance in Appendix A.2.

Building upon the observation that the exploration for optimal cross-modal interaction among visual and textual representations remains a yet-to-be-explored challenge in temporal grounding, our Correlation-Guided DEtection TRansformer~(CG-DETR) employs a novel paradigm of adaptive textual conditioning, enabling the model to attend more to desired moments. 
% on discovering the appropriate degree of interaction and utilizing such degrees in predictions.
% Building upon the observation that the exploration for optimal cross-modal interaction among visual and texture representations remains a yet-to-be-explored challenge in temporal grounding, our Correlation-Guided DEtection TRansformer~(CG-DETR) focuses on discovering the appropriate degree of interaction and utilizing such degrees in predictions.
To this end, we first propose an Adaptive Cross-Attention layer~(ACA) that modifies the operation of the cross-attention layer; by adding dummy tokens to the {\it key} in the multi-head attention, we adjust the intensity of text query's clip-wise engagement based on the ground truth relevance between video clips and a text query.
Subsequently, we delve into the relationship between clips and each word in a text query.
It is evident that not all words bear close relevance to a video clip, even if they are part of a highly relevant text query.
Still, studying a fine-grained correlation is demanding since word-level supervision requires huge resources.
% Still, studying a fine-grained correlation is demanding since preparing supervision for each word requires huge resources.
Hence, we estimate the relevance between individual words and video clips by computing their similarity within a modality-aligned space in the broader concept that considers the overall sentence and video context.
% Thus, we approximate the clip-word relevance by calculating similarity in a modality-aligned space at a more coarse video and sentence level.
The similarity map is then, distilled to adjust the magnitude of the attention map in the cross-attention. 
Lastly, we propose a moment-adaptive saliency detector that leverages instance-specific context and the magnitude of the cross-modal interaction whereas previous saliency detectors rely on learnable parameters.
A saliency detector, after being encoded instance-specifically with each video-text representation, is utilized to calculate the saliency scores, \textit{i.e.,} the highlightness, by computing its similarity to each video clip.
% A saliency token designed with two principles, \textit{i.e.}, preserving original context and representing diverse characteristics of desired moments, is aggregated with textual information.
This allows us to incorporate scaled interaction degrees into prediction.

To sum up, our contributions are (i)~We propose adaptive cross-attention with dummy tokens to enable the manipulation in the degree of video-text interaction with respect to their correlation, (ii)~To further calibrate the interaction degree, we discover and distill the clip-word correlation from the aligned space at video-sentence level, (iii)~We introduce a moment-adaptive saliency detector to exploit the calibrated degree of cross-modal interaction, and (iv) Extensive experiments demonstrate the effectiveness of CG-DETR.
\vspace{-0.1cm}
\section{Related Work}
\vspace{-0.1cm}
% \subsection{Moment Retrieval and Video Highlight Detection}
% \subsection{Video Temporal Grounding}
%\textbf{Video temporal grounding}, a task to link the text query to corresponding video segments, can be further divided into moment retrieval and highlight detection which localizes the desired moments and scores the clip-wise correspondence to the query.
%Moment retrieval has been introduced with the goal of retrieving user-desired moments~\cite{anne2017localizing, gao2017tall, soldan2021vlg, mr1, mr2, mr3, mr4}.
Video temporal grounding, which involves matching text queries to specific video segments, can be categorized into moment retrieval and highlight detection. Moment retrieval identifies the segments relevant user-specified text query~\cite{anne2017localizing, gao2017tall, soldan2021vlg, mr1, mr2, mr3, mr4}.
Conventional approaches fall into either proposal-based or proposal-free methods.
Proposal-based methods employ predefined proposals, \textit{e.g.}, sliding windows~\cite{anne2017localizing, liu2018attentive, zhang2019exploiting, ge2019mac, yuan2019semantic} and temporal anchors~\cite{yuan2019semantic, chen2018temporally,zhang2019cross,zhang2019man,liu2020jointly}, or learn to generate proposals~\cite{zhang2020learning,liu2021context,xiao2021boundary,shao2018find,xu2019multilevel} to prepare moment candidates. 
With the candidates, they treat the task as the matching problem between candidates and the text query. 
On the other hand, proposal-free methods learn to encode multimodal knowledge and predict the temporal spans with the regression head.
Highlight detection aims to score the importance of every clip with respect to either visual or textual information.
It has been studied extensively with varying sources, \textit{e.g.}, visual only~\cite{rochan2020adaptive, hl4, badamdorj2022contrastive, xiong2019less} and visual-audio~\cite{xiong2019less, hl1, hl2, hl5}, and different granularity levels on the given labels; there exist supervised~\cite{hl4, gygli2016video2gif, youtubehl}, weakly supervised~\cite{xiong2019less, cai2018weakly, panda2017weakly}, and unsupervised methods~\cite{badamdorj2022contrastive, mahasseni2017unsupervised, khosla2013large, rochan2018video} for highlight detection.
% It has been studied extensively with varying sources, \textit{e.g.}, visual only and visual-audio~\cite{hl1, hl2, hl3, hl4, hl5}, and different granularity levels on the given labels; highlight detection methods can be classified into supervised~\cite{gygli2016video2gif, xu2021cross, youtubehl}, weakly supervised~\cite{cai2018weakly, panda2017weakly, xiong2019less}, and unsupervised methods~\cite{badamdorj2022contrastive, mahasseni2017unsupervised, khosla2013large, rochan2018video}.

Since the advent of QVHighlights~\cite{momentdetr}, these have been addressed together. % and are known to be mutually beneficial~\cite{umt}.
% It was not until recently that these problems have been handled concurrently.
% With the introduction of QVHighlights~\cite{momentdetr}, a dataset for joint moment retrieval and highlight detection, it became evident that addressing these tasks together is mutually beneficial~\cite{umt}.
Recent approaches can be divided into DETR- or regression-based.
Employing either stream, diverse practices were conducted; UMT~\cite{umt} exploited additional audio modality, QD-DETR~\cite{qddetr} and EaTR~\cite{eatr} developed the DETR architecture, and \cite{univtg, unloc} remarked the importance of pretraining.
Also, the study for the query dependency is being spotlighted~\cite{qddetr, hou2020conquer, barrios2023localizing, hendricks-etal-2018-localizing, liu2023survey}.
Our motivation resembles these works, especially QD-DETR~\cite{qddetr}.
% However, whereas they only focus on enforcing text engagement in every video clip, we aim to discover and calibrate to an appropriate degree of query dependency in cross-modal interaction.
However, whereas they only focus on enforcing text engagement in every video clip, we aim to discover an appropriate degree and calibrate the query dependency in cross-modal interaction.
More related works about vision-text alignment and hierarchical interaction are in Appendix A.1.

\begin{figure*}[t!]
    \centering
    \vspace{-0.6cm}
    \includegraphics[width=1.\textwidth]{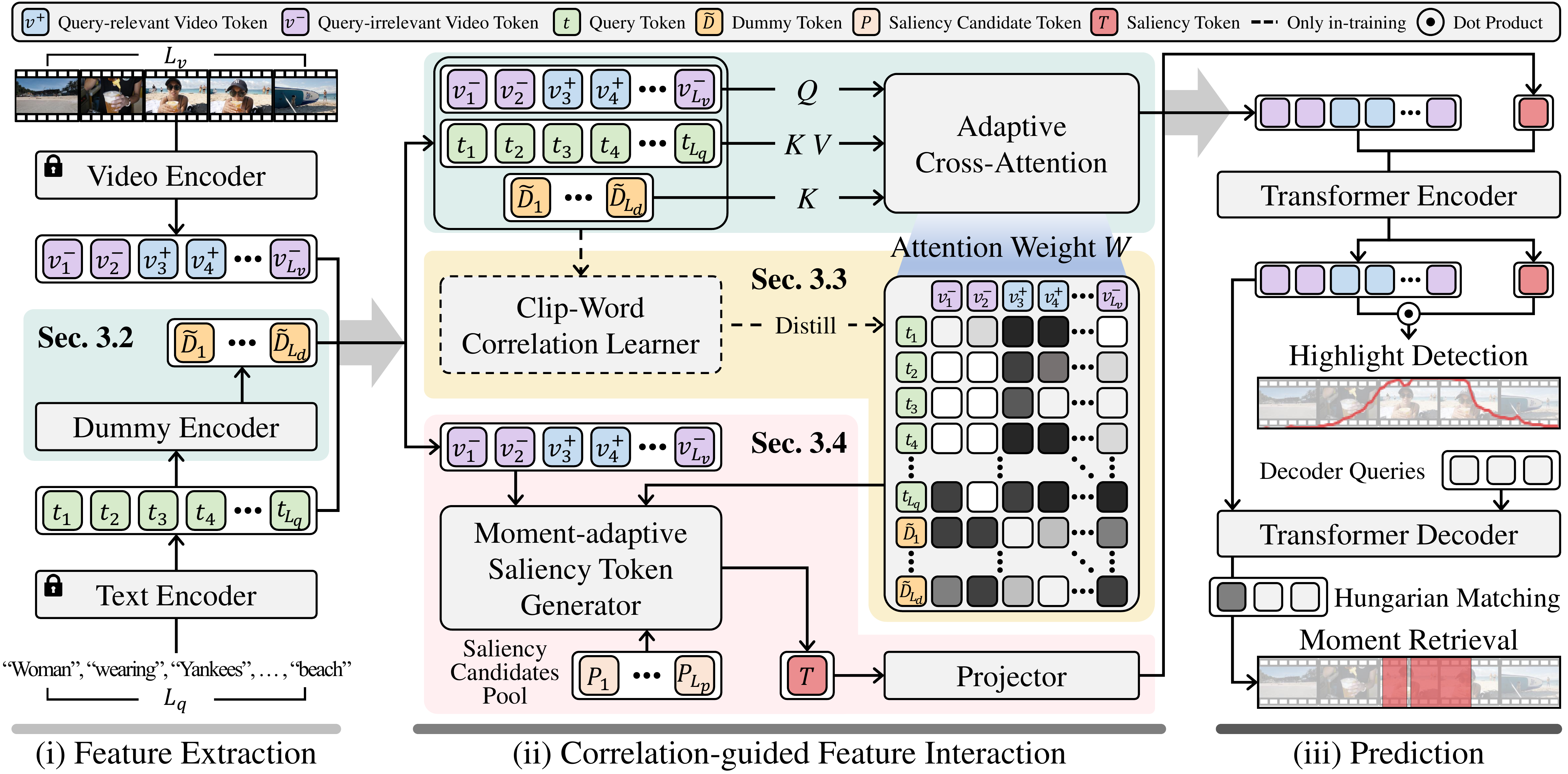}
    \vspace{-0.6cm}
    \caption{CG-DETR overview. From left to right, the model consists of three phases: (i)~feature extraction, (ii)~correlation-guided feature interaction, and (iii)~predictions for grounding tasks.
    (i) Along with video and text feature extraction, dummy tokens are conditioned by the query to represent the query-excluding meaning.
    (ii) Correlation-guided feature interaction is performed with adaptive cross-attention. 
    In addition to calibration of text query engagement as a whole, we also guide the word-wise engagement with clip-word correlation learner.
    % Not only do we calibrate the text query engagement as a whole, but we also guide the word-wise engagement with the correlation learner.
    At the bottom, a saliency token $T$ is generated with video tokens and saliency candidates according to the value of calibrated attention map.
    A saliency token is processed via a projector that shares the parameters with {\it query} $\textit{Q}$ projection layer in adaptive cross-attention.
    % At the bottom, we utilize the video context and a calibrated attention map to generate a saliency token. 
    % A saliency token $T$ is processed via \textit{query} projection layer in the adaptive cross-attention to prevent a modality gap with the video tokens.
    % A saliency token is processed through the same cross-attention layer to be similarly encoded with video tokens to prevent a modality gap. 
    Details for correlation learner and saliency token are in Fig.~\ref{fig:SCM}, ~\ref{fig:MSD}.
    (iii) Finally, tokens are processed through the encoder and decoder to make predictions.
    % \SE{}
    % 1) i, ii, iii 에 대한 표기가 그림에도 포함되면 더 좋을거같아요
    % 2) 두번째 문장이 조금 더 자세하면 좋을거 같기는 합니다. (dummy token에 대한 설명 / aca에서 attention weight가 guide 된다는 점? / guide를 위한 correlation이 어떤건지 (video-setenec aligned space?))
    % 다른 figure들 이랑 같이봐야할거 같긴한데... 이게 제일 먼저 나오는거니까 반복되더라도 조금 더 자세한 설명은 포함되는게 좋지 않나 싶어요
     }
    \label{fig:main_figure}
    \vspace{-0.6cm}
\end{figure*}

% mining / generator 는 뒤에있다고 설명
% \vspace{0.1cm}
\vspace{-0.2cm}
\section{Method}
\vspace{-0.1cm}
% \subsection{Overview} % section 어디어디에 각각 있는지
Given a video $V = [v_1, v_2, ..., v_{L_v}]$ of $L_v$ clips and a text query $Q = [q_1, q_2, ..., q_{L_q}]$ of $L_q$ tokens, our objective is to predict clip-wise saliency scores $\{s_1, s_2, ..., s_{L_v}\}$ and to localize the target moments in the form of $(m_c, m_\sigma)$ where $m_c$ and $m_\sigma$ denote the center temporal coordinate and span of the moment.
The overview of our model is illustrated in Fig.~\ref{fig:main_figure}.

The aim of CG-DETR is to leverage the interrelation between modalities to improve the feature interaction operation.
To achieve that, in Sec.~\ref{aca}, we propose to employ the dummy tokens $D$; these serve to distinguish the degree of cross-modal attention for each video clip and are conditioned by query tokens to encapsulate representations that exclude query contexts.
In Sec.~\ref{sec.3.3}, we provide fine-grained correlation guidance to the adaptive cross-attention layers by aligning the modalities at the video-sentence level where supervision is available and inferring at a more granular clip-word level. 
Lastly, in Sec.~\ref{sec.3.4}, we introduce a saliency token that incorporates both video context and moment-adaptive knowledge derived from clip-wise query correspondence in the learned attention map of cross-attention layers.
% Then, a saliency token $T$ is processed through the projection layers that share parameters with the one in the adaptive cross-attention to prevent be forwarded for the prediction.
Then, a saliency token $T$ is processed through the projection layer to fit the output space of the adaptive cross-attention.
% in the adaptive cross-attention to be forwarded for the prediction.
% Then, a saliency token processed via cross-attention exclusively with text tokens is concatenated with clip-wise tokens to be forwarded for the prediction.

% 먼저 video /text feature를 뽑고
% interaction을 담당하는 부분,
% encoder, decoder를 거쳐서 

\subsection{Adaptive Cross-Attention~(ACA): Reflecting the Video-Text Correlation}
\label{aca}
\vspace{-0.1cm}
Adaptive cross-attention stems from the insight that not all contents in a video correspond to the semantics of a text query.
Meanwhile, cross-attention is one of the most popular methods to incorporate the features of different modalities, especially for retrieval tasks~\cite{qddetr, song2021image}.
Yet, as the softmax applied in the cross-attention forces the text query to be equally aggregated across all video clips, the cross-attention is inadequate to learn the degree of video-text correspondence. 
A na\"ive solution is to change the softmax activation to sigmoid.
However, we point out the vulnerability of sigmoid in ranking the text-relevance scores of clips since it eliminates the dependency among other clips.
% A na\"ive solution is to change the softmax activation to sigmoid.
% However, we point out the vulnerability of sigmoid in ranking the text-relevance scores of clips since it eliminates the dependency among other clips.

% However, it exposes vulnerability to rank the text-relevance scores of each clip, \textit{i.e.}, highlight detection, since the sigmoid activation disposes the dependency to similarity scores between other clips and text query.

% To design the adaptive cross-attention, the na\"ive implementation would be changing the activation function, \textit{i.e.}, softmax to sigmoid.
% However, it exposes vulnerability to rank the text-relevance scores of each clip, \textit{i.e.}, highlight detection, since the sigmoid activation disposes the dependency to similarity scores between other clips and text query.

% \begin{figure}[t!]
%     \centering
%     \vspace{-0.4cm}
%     \includegraphics[width=0.5\columnwidth]{figure/fig_3.pdf}
%     \vspace{-0.3cm}
%     \caption{Illustration of deriving clip-wise query correspondence $\bar{a}$.
%     }
%     \vspace{-0.5cm}
%     \label{fig:textattendance}
% \end{figure}
\begin{wrapfigure}{t!}{7.2cm}
    \centering
    \vspace{-0.55cm}
    \includegraphics[width=0.5\columnwidth]{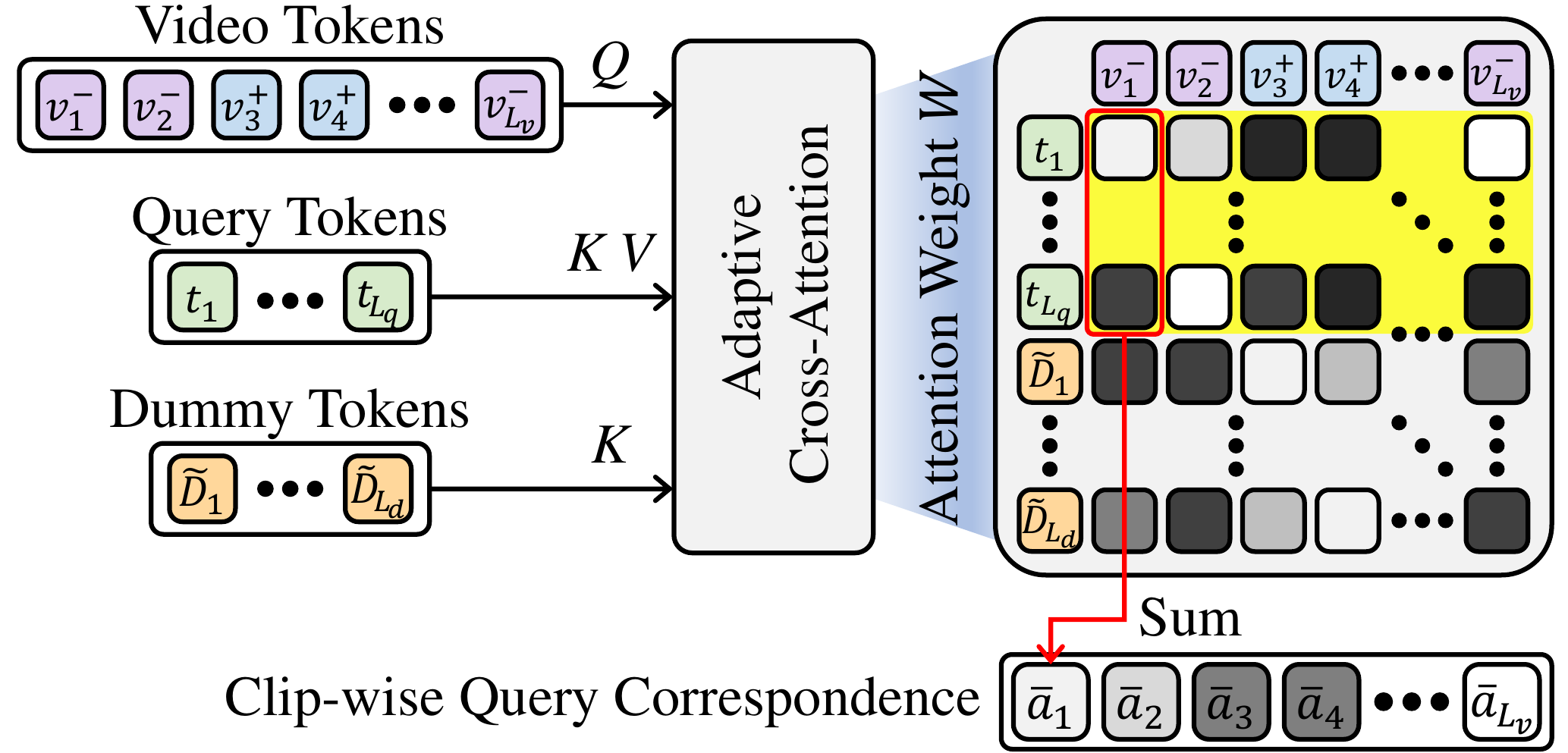}
    \vspace{-0.1cm}
    \caption{Illustration of deriving clip-wise query correspondence $\bar{a}$.
    }
    \vspace{-0.4cm}
    \label{fig:textattendance}
\end{wrapfigure} 

Instead, we introduce dummy tokens that are concatenated to the text tokens and possess a portion of attention weights in cross-attention layers~\cite{liu2021progressively}.
Yet, employing universally shared dummies without specific roles may not effectively distribute attention weights in cross-attention.
To address this, we encode instance-adaptive dummies and explicitly guide dummies to occupy the attention weights inversely proportional to the text-visual relevance.
Assuming that we have $L_d$ learnable dummy tokens $D = [D_1, D_2, ..., D_{L_d}]$, these dummy tokens are conditioned by the text query $Q$ through attention layers to contain the query-excluding context.
We denote the encoded dummy tokens as $\tilde{D}$. 
% Dummy tokens $D = [D_1, ..., D_k]$, defined as K learnable vectors, are first projected with text query $Q$ to contain the opposite context to $Q$.
Then, in the cross-attention layer where {\it query} $\textit{Q} = [p_\textit{Q}(v_1), p_\textit{Q}(v_2), ..., p_\textit{Q}(v_{L_v})]$ and {\it value} $\textit{V} = [p_\textit{V}(t_1), p_\textit{V}(t_2), ..., p_\textit{V}(t_{L_q})]$ are prepared by projecting the video clips and text queries, dummy tokens take part in {\it key} $\textit{K} = [p_\textit{K}(t_1), p_\textit{K}(t_2), ..., p_\textit{K}(t_{L_q}), p_\textit{K}(\tilde{D}_1), p_\textit{K}(\tilde{D}_2), ..., p_\textit{K}(\tilde{D}_{L_d})]$, concatenated with text tokens before {\it key} projection. 
$p_\textit{Q}(\cdot)$, $p_\textit{K}(\cdot)$, and $p_\textit{V}(\cdot)$ are projection layers for {\it query}, {\it key}, and {\it value}.
% Formally, the operation of the cross-attention layer with dummy tokens is expressed as:
% \begin{eqnarray}
%     \label{eqn 1 cross attention}
%     \text{VAR1} = \{{\textit{Q}\textit{K}^T}_{i,j} \mid 0 \leq i < L_v, 0 \leq j < L_q \}\\
%     \text{Attention}(\textit{Q}, \textit{K}, \textit{V}) = \text{softmax}(\frac{\text{VAR1}}{\sqrt{d}})\textit{V}_t,
% \end{eqnarray}
Formally, the operation of adaptive cross-attention with dummy tokens for $i$-th video clip, $\text{ACA}(v_i)$, is expressed as:
\vspace{-0.3cm}
\begin{eqnarray}
    \label{eqn 1 cross attention}
    \text{ACA}(v_i)\!=\!\sum_{j=1}^{L_q}W_{i,j} \!\odot\! \textit{V}_j; 
    \; W_{i,j}\!= \frac{
        {\text{exp}\left( \frac{\textit{Q}_i \odot \textit{K}_j} {\sqrt{h}} \right)}}
        {\sum_{k=1}^{L_q+L_d}\!\text{exp}\!\left( \frac{\textit{Q}_i \odot \textit{K}_k}{\sqrt{h}} \right)},\!\!\!
\end{eqnarray}
% \vspace{-0.1cm}
where $h$ denotes the projected hidden dimension and $\odot$ stands for the dot product.
% To guide the dummy tokens to take desired portions of attention weights, we calculate clip-wise query correspondence scores $\hat{a}$ as shown in Fig.~\ref{fig:textattendance} and apply objectives for highlight detection.
To guide the dummy tokens to take desired portions of attention weights that are inversely proportional to the saliency scores, we define a query correspondence score for $i$-th clip, $\bar{a}_i=\sum_{j=1}^{L_q}W_{i,j}$, as shown in Fig.~\ref{fig:textattendance} and train $\bar{a}_i$ by applying the objectives for highlight detection.
% calculate clip-wise query correspondence scores $\bar{a}$ by summation over query words for a given clip as shown in Fig.~\ref{fig:textattendance} and apply objectives for highlight detection.
% Formally, $\bar{a}$ can be expressed as $\bar{a_i}=\sum_{j=1}^{L_q}W_{i,j}$.
% with the attention weights; we calculate clip-wise correspondence scores $\hat{a}$ as shown in Fig.~\ref{fig:textattendance} and apply the loss on $\hat{a}$.
% by adding the text attendance for each video clip to serve as saliency scores.
Details for the objective are in Sec.~\ref{sec.3.5} and Appendix A.5.
In addition, we employ binary cross-entropy to discretize between moment and non-moments within every instance and enforce orthogonality among encoded dummy tokens to prevent them from playing the same role:
% To prevent the dummy tokens from playing the same roles, we also enforce orthogonality between output dummy tokens as:
% \begin{align}
%     \label{eqn dummy_orthogonal}
%     & \mathcal{L}_{bce}=\frac{1}{L_v}\sum^{L_v}_{i=1}\left(a \odot \text{log} \;\hat{a} + \left(1-a\right) \odot \text{log}\left(1-\hat{a}\right)\right), \\
%     & \mathcal{L}_{ortho}=\frac{1}{k\left(k-1\right)}\sum^{k}_{i=1}{\sum^{k}_{j=1}{\mathbbm{1}_{i\neq j} |\hat{D}_i\odot \hat{D}_j|}},
% \end{align}
% \SB{} % \hat{a}에 대한 정의... 이게 필요한가......
% \begin{equation}
%     \label{query correspondence score}
%     \hat{a_i}=\sum_{j=1}^{L_q}W_{i,j}
% \end{equation}
\vspace{-0.2cm}
\begin{equation}
    \label{eqn dummy_orthogonal}
    % \mathcal{L}_{\text{bce}}=a \odot \text{log} \;\hat{a} + \left(1-a\right) \odot \text{log}\left(1-\hat{a}\right),
    \mathcal{L}_{\text{bce}}=\!\frac{1}{L_v}\!\sum^{L_v}_{i=1}\!\left(a_i \!\odot\! \text{log} \;\bar{a}_i + \left(1-a_i\right) \!\odot\! \text{log}\left(1-\bar{a}_i\right)\right),
\end{equation}
\vspace{-0.5cm}
\begin{equation}
    \mathcal{L}_{\text{ortho}}=\frac{1}{L_d\left(L_d-1\right)}\sum^{L_d}_{m=1}{\sum^{L_d}_{n=1}{\mathbbm{1}_{m\neq n} |\tilde{D}_m\odot \tilde{D}_n|}},
% dummy 갯수 k대신 다른걸로 표기
\end{equation}
where $a_i$ is 1 if saliency GT~(text-video relevance) is non-zero, otherwise set to 0.
By allowing such flexible cross-attention, the model learns better separation of relevant and irrelevant video segments.

% \SE{} % 여기 딱 읽고 잠깐 든 생각 중 하나가, SA 같은 경우는 이것과 같은 guidance (HD objective + BCE 등)은 줄 수 있을텐데 그것 대비 나은점도 작성하면 어떤가 싶네요, 실험도 있으면? 간단하게 생각했을 땐, txt를 없앤다고 vid의 영향력을 키우는건 MR자체에 악영향을 가져올 수 있다 이런 내용을 적을 수 있을까 싶어요

% Instead, we concatenate the dummy token to each input text that can possess a portion of attention weights in conventional cross-attention layers.
% Formally,  
% where a dummy token is induced by another attention mechanism with input text to take semantic context into account.

% More Flexible Relevance Detection: In many real-world scenarios, not all video clips will be directly relevant to the given text, and vice versa. By allowing the attention mechanism to assign lower attention weights or relevance scores to certain tokens or video segments when they are not relevant to each other, you are making the model more flexible in detecting varying degrees of relevance. This is important because the relevance between text and video moments can be nuanced, and not all moments may have the same level of correspondence to the text.

% Enhanced Discrimination: The modified approach allows the model to better discriminate between relevant and irrelevant moments by providing the freedom to assign lower attention values when necessary. This can lead to better separation of relevant and irrelevant video segments, which is crucial for accurate video moment retrieval.

% High level align
% Low level discover and reflect

\subsection{Clip-Word Correlation Learner: Align, Discover, and Distill the Fine-grained Correlation}
\label{sec.3.3}
\vspace{-0.1cm}
% \subsection{Align, Discover, and Distill:\\ Delving into the Fine-grained Correlation}
% \subsection{Delving into fine-grained correlation:\\ Align, Discover, and Distill}
% \subsection{Align, Discover, and Reflect the Clip-Word Correlation}
% Fine Distillation: Discover and Reflect the Clip-Word Correlation}
% \subsection{Self-supervised Correlation Mining: Discover and Reflect the Fine-grained Correlation}
% \subsection{Fine-grained Correlation Discover and Distill:\\ Reflecting Fine-grained Correlation}
% \subsection{Self-supervised Modality Alignment: \\Guiding the cross-attention weights}

% \SB{} % 뒤에서 사용될지도 모르는 수식들...
% \begin{equation}
%     V^{b+}=\left[v^{b}_i | i\in\{1, 2, \dots, L_q\}, a_i=1\right],
% \end{equation}
% \begin{equation}
%     V^{b-}=\left[v^{b}_i | i\in\{1, 2, \dots, L_q\}, a_i\neq1\right],
% \end{equation}

By employing ACA, the model considers the clip-wise relevance with text query.
Delving into the clip-word relation, it is evident that words are not equally associated with visual clues.
With this motivation, we further aim to discover the appropriate attention weights for cross-attention that represent the fine-grained correlation between video clips and words.

\vspace{-0.0cm}

% \vspace{-0.4cm}
\begin{figure}[t!]
    \centering
    \vspace{-0.2cm}
    \includegraphics[width=1.\columnwidth]{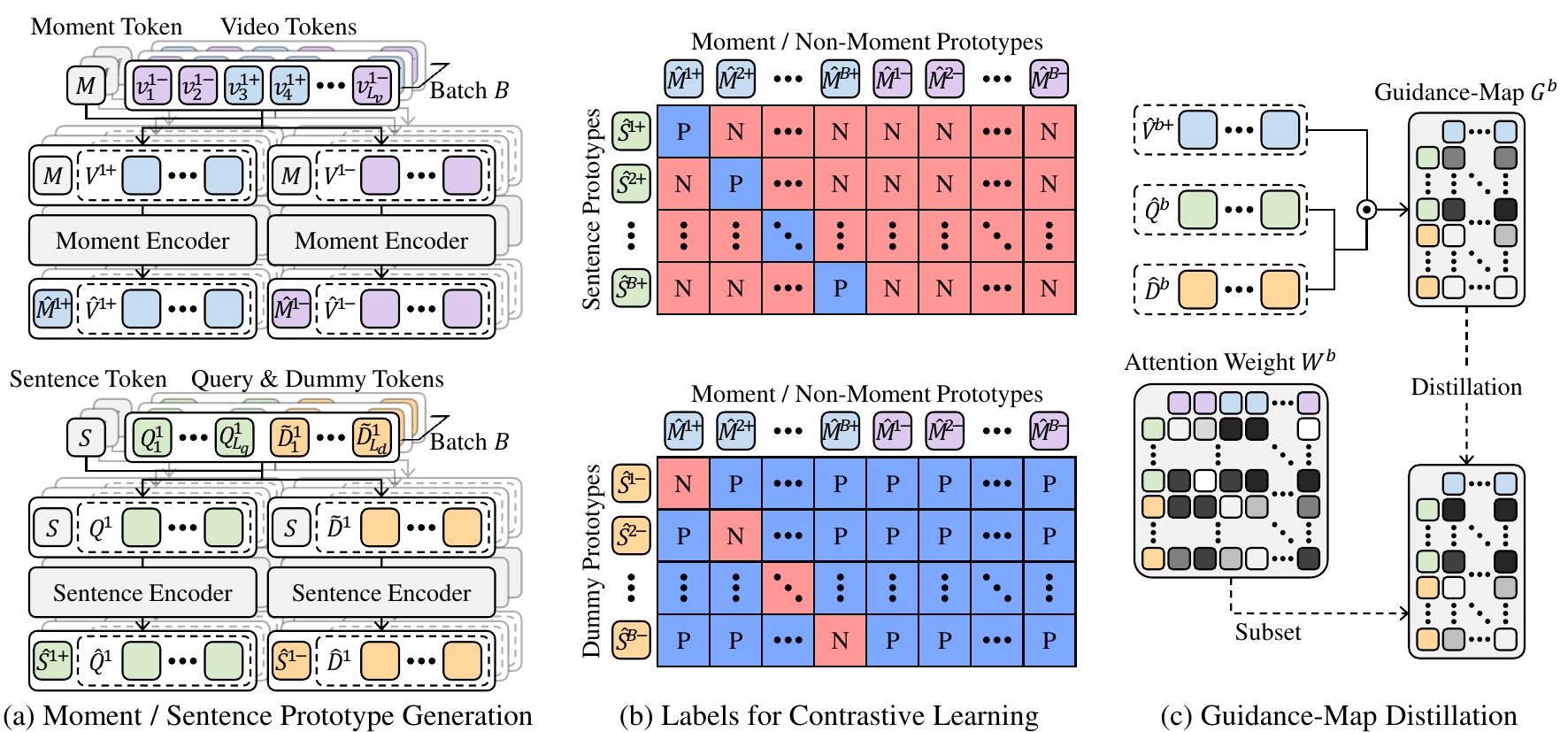}
    \vspace{-0.3cm}
    \caption{
    % Clip-word correlation learner. We discover and reflect the relevance between clips and text words into cross-attention.
    Clip-word correlation learner to reflect the relevance between clips and text words into cross-attention.
    % relevance between clips and text words
    (a) We establish visual moment, non-moment, query, and dummy prototype tokens~($\hat{M}^{b+}, \hat{M}^{b-}, \hat{S}^{b+}, \text{and } \hat{S}^{b-}$ of $b$-th instance within a batch) using learnable moment and sentence tokens.
    (b) To learn the aligned space, we use contrastive learning. Whereas the query prototype $\hat{S}^{+}$ learns to be aligned with the paired visual moment token $\hat{M}^{+}$, dummy prototype $\hat{S}^{-}$ learn to exclude the moment-specific knowledge.
    (c) Given the moment-sentence aligned space from (b), we infer the correlation in clip-word level between each clip and text words as well as dummy tokens to form guidance map $G$. 
    Then, guidance is provided to the attention map in the cross-attention. 
    }
    \label{fig:SCM}
    \vspace{-0.5cm}
\end{figure}

% 원본
However, determining the proper clip-word correlation presents a significant challenge in the absence of direct supervision.
To address this, we take a two-step approach; 1) learning the video-text-aligned embedding space at the moment-sentence level, where supervision is available, 2) inferring the video-text correlation at a more fine-grained level, \textit{i.e.}, clip-word level, and distilling the correlation into the attention maps within the cross-attention layers.
This methodology enables the moment-sentence pair information to be further utilized as a form of weak supervision for learning fine-grained correlation.
% In short, we aim to support the multimodal interaction in clip-word level with the given weak supervision 
% \SE{} % 모르겠다.... 뭐라 적을지 모르겟어요ㅣ...
% However, determining the proper clip-word correlation is challenging in the absence of supervision.
% Hence, we first aim to learn the vision-text-aligned embedding space at the video-query level where moment-query supervision is given. --> moment-query 라고하니까 조금 그래서 위에 video-text pair onfo 추ㅏㄱ.
% This process makes the moment embedding similar to the matched text query, and simultaneously, to be dissimilar to the corresponding dummy tokens. --> dummy만이아니라 다른 비디오랑도 그래서 다시 원본에서 수정..
% Then, based on the aligned space, we infer the correlation between clips and words and dummies. 
% As this correlation has a proper vision-text relevance, we distill the correlation map to the attention map in the cross-attention layers.

% To align the embedding space of both modalities at a video-query level, we use contrastive learning~\cite{chen2020simple, CLIP, groundingdino, GLIP}.
To align the embedding of both modalities at a video-sentence level, we generate moment and non-moment prototypes for both modality domains with domain-specific prototype tokens, \textit{i.e.}, moment and sentence tokens, and apply contrastive learning~\cite{chen2020simple, CLIP, groundingdino, GLIP} between them.
% ~\cite{chen2020simple, CLIP, groundingdino, GLIP}
Let $V^{b+}$ $=$ $\left[v^{b}_i | i\in\{1, 2, ..., L_v\}, a_i=1\right]$ and $V^{b-}$ $=$ $\left[v^{b}_i | i\in\{1, 2, ..., L_v\}, a_i\neq1\right]$ be the video clips in the specified target moment~(query-relevant) and otherwise~(query-irrelevant) in $b$-th video instance of batch size $B$, respectively.
$a$ indicates the GT for query correspondence score $\bar{a}$, as mentioned in Sec.~\ref{aca}.
In Fig.~\ref{fig:SCM}~(a), we illustrate prototype generation.
By processing learnable moment token $M$ with each of these visual inputs through the self-attention block~(SA), we derive moment prototype $\hat{M}^{b+}$ and non-moment prototypes $\hat{M}^{b-}$ for $b$-th video instance in the visual domain each with the projected video representations $\hat{V}^{b+}$ and $\hat{V}^{b-}$:
% By processing learnable moment token $M$ with each of these inputs through the self-attention block~(SA), we derive visual pro$b$-th instance-specific visual moment prototype $\hat{M}^{b+}$ and non-moment prototype $\hat{M}^{b-}$ with projected video representations $\hat{V}^{b+}$ and $\hat{V}^{b-}$:
\begin{align}
    \hat{M}^{b+}, \hat{V}^{b+}\!\! =\text{SA}([M; V^{b+}] )\;,\;\;
    % \\
    \hat{M}^{b-}, \hat{V}^{b-}\!\! =\text{SA}([M; V^{b-}] ),
\end{align}
% \endgroup
% \vspace{-0.cm}
% \begin{equation}
%     \hat{M}^{b+}, \hat{V}^{b+} =\text{SA}([M; V^{b+}], [M; V^{b+}], [M; V^{b+}] ),
% \end{equation}
% \vspace{-0.65cm}
% % \begingroup
% % \setlength{\belowdisplayskip}{-0.01pt}
% \begin{equation}
%     \hat{M}^{b-}, \hat{V}^{b-} =\text{SA}([M; V^{b-}], [M; V^{b-}], [M; V^{b-}] ),
% \end{equation}
% % \endgroup
where $\left[~;\right]$ and $\;\hat{\cdot}\;$ denote concatenation and projected output via SA, respectively.
% where $\left[~;\right]$ and $\;\hat{\alpha}\;$ denote concatenation and projected version of $\alpha$ via SA, respectively.
Symmetrical to processing the moment token $M$ in the video domain, we leverage the learnable sentence token $S$ with each of the $b$-th text queries $Q^b$ and dummy tokens $D^b$ to yield moment and non-moment prototypes in the semantic domain:
\vspace{-0.15cm}
% Similarly, leveraging the $b$-th text query $Q^b$ and dummy tokens $D^b$ in conjunction with the learnable sentence token $S$, we construct moment and non-moment prototypes in the semantic domain:
% from the perspective of the query and the dummy:
\begin{align}
    \hat{S}^{b+},\hat{Q}^b=\text{SA}([S; Q^b ]) \; , \;\;
    \hat{S}^{b-},\hat{D}^b=\text{SA}([S; \tilde{D}^b ] ).
\end{align}
% \vspace{-0.1cm}
% \begin{equation}
%     \hat{S}^{b+},\hat{Q}^b=\text{SA}([S; Q^b ], [S; Q^b ], [S; Q^b ] ),
% \end{equation}
% \vspace{-0.2cm}
% \begin{equation}
%     \hat{S}^{b-},\hat{D}^b=\text{SA}([ S; \tilde{D}^b ], [S; \tilde{D}^b ], [S; \tilde{D}^b ] ).
% \end{equation}
% These prototypes are referred to as $\hat{S}^{b+}$ and $\hat{S}^{b-}$.
% For future reference, we also introduce $\hat{V}^{b+}$ and $\hat{V}^{b-}$ to represent projected versions of $V^{b+}$ and $V^{b-}$ using the same projection layer employed in generating $\hat{M}$; $\hat{Q}$ and $\hat{D}$ are projected $Q$ and $D$ through attention block with $S$, respectively.

Consequently, with $\mathcal{B}$ as the index set in the batch, we formulate the contrastive learning objective for $b$-th instance using the batch-wise labels in Fig.~\ref{fig:SCM}~(b):
\vspace{-0.2cm}
\begin{gather}
    \mathcal{L}^+_{\text{align}}\!=\! -\text{log}\frac{\text{exp}(\hat{S}^{b+} \odot \hat{M}^{b+} / \tau )}
    {\sum\limits_{o\in \mathcal{B}}\sum\limits_{* \in \left\{+,-\right\}}\text{exp}(\hat{S}^{b+} \odot \hat{M}^{o*} / \tau )},
    \\
    \mathcal{L}^-_{\text{align}}=
    -\text{log}\bigg(1-\frac{\text{exp}(\hat{S}^{b-} \odot \hat{M}^{b-} / \tau )}
    {\sum\limits_{o \in \mathcal{B}}\sum\limits_{* \in \left\{+,-\right\}}\!\!\!\!\!\text{exp}(\hat{S}^{b-}\! \odot \!\hat{M}^{o*} / \tau )}\bigg),
    \\
    \mathcal{L}_{\text{align}} = \mathcal{L}^+_{\text{align}} + \mathcal{L}^-_{\text{align}}.
\end{gather}
To illustrate, $\mathcal{L}^+_{\text{align}}$ is to enforce the proximity between the visual moment prototypes $\hat{M}^{b+}$ and semantic moment prototypes $\hat{S}^{b+}$ whereas $\mathcal{L}^-_{\text{align}}$ is to assign the moment-excluding visual representation in the dummy tokens.
The rationale behind the large semantic coverage in the dummy tokens stems from the challenge of defining clear antonyms within the complex, high-dimensional spaces of moments and sentences.
% The rationale behind the large semantic coverage in the dummy tokens is due to the obscureness of the opposite meaning in terms of high-dimensional moment and sentence space.
For instance, given that a sentence consists of words, each with potential antonyms, explicitly defining precise opposite representations is difficult.
% For instance, the sentence consists of words and there exist opposite terms for every word which makes it hard to explicitly define the opposite representation in the high-dimensional sentence space.
Hence, we instead train the dummy tokens to encompass a wide range of semantics, excluding only the particular semantics.
% For instance, the opposite term may exist for every word that consists of the sentence which makes it hard to define the opposite definition in high-dimensional space.

% The design choice of the role of dummy tokens for why they have such large coverage in representing visual space is due to the obscureness of the opposite meaning in terms of high-dimensional moment and query space.
% The design choice of the role of dummy tokens is due to the obscureness of the opposite meaning in terms of high-dimensional moment and query space.

% 3. Sentence
% Positive set -> Pair positive moment
% Negative set -> Non-Paired Positive moment & All Negative Moment

% 4. Dummy Sentence
% Positive set -> Non-Paired Positive moment & All Negative Moment
% Negative set -> Paired positive moment
% 출처 : A \euiqv I 이거는 supervised contrastive learning에서 따온 수식.

Then, we infer the word-clip correlation from the aligned space to derive the normalized guidance $G$~(Fig.~\ref{fig:SCM}~(c)) for the attention map in the adaptive cross-attention~(Normalizing the guidance isolates the relative correlation between words from the overall relevance between the whole text query and the video which is determined using ACA):
% \begin{eqnarray}
%     \label{guidance}
%     G_{i,j} = \hat{v}^{+}_i \odot {\left(\left[\hat{Q}_i, \hat{D}_i\right]\right)^{T}},
% \end{eqnarray}
\vspace{-0.15cm}
% \begingroup
% \setlength{\belowdisplayskip}{0.001pt}
\begin{eqnarray}
    \label{guidance}
    G_{i,j} = \frac{\text{exp}\left(\hat{v}^{+}_{i} \odot {[\hat{Q}; \hat{D}]}_j \right)}{\sum_{k=1}^{L_q+L_d}{\text{exp}\left(\hat{v}^{+}_{i} \odot {[\hat{Q}; \hat{D}]}_k \right)}},
% dummy k 대신 다른거
\end{eqnarray}
% \endgroup
% \vspace{0.1cm}
where $\hat{v}^{+}_i$ is $i$-th video clip of $\hat{V}^+$.
Note that we only derive the guidance for positive clips, \textit{i.e.}, clips belonging to the GT moments, since the model has not learned the positive textual relationships for non-moment clips which can make it susceptible to inaccuracies.
% by calculating the correlation between the clip features belonging to the ground-truth moments $\hat{v}^+_i$ and the word-wise text features: 
% \begin{eqnarray}
%     \label{guidance}
%     G_i = \hat{v}^{+}_i \odot {\left(\hat{Q}_i \concatA \hat{D}_i\right)^{T}},
% \end{eqnarray}
% \begin{eqnarray}
%     \label{guidance}
%     G_{i,j} = \hat{v}^{+}_i \odot {\left(\hat{Q}_i \concatA \hat{D}_i\right)^{T}},
% \end{eqnarray}
% where $\concatA$ denotes concatenation operation.
Finally, we provide clip-word level guidance to the attention map in the cross-attention layer.
% Given the weights in the attention map $W$ and the guidance $G$, distillation loss is expressed as:
Given the weights in the attention map $W$ defined in Eq.~\ref{eqn 1 cross attention} and the guidance $G$ defined in Eq.~\ref{guidance}, distillation loss $\mathcal{L}_{\text{distill}}$ is expressed as: 
\vspace{-0.15cm}
% \begingroup
% \setlength{\belowdisplayskip}{-0.0001pt}
\begin{eqnarray}
    \label{eqn distill loss} 
    \mathcal{L}_{\text{distill}} = \frac{1}{L_v}\sum_{i=1}^{L_v}\sum^{L_q + L_d}_{j=1}\mathbbm{1}_{{v_i}\in{V^+}}W_{i,j}~\text{log}\frac{W_{i,j}}{G_{i,j}}.
\end{eqnarray}
% \endgroup
% \vspace{0.1cm}
% \abovedisplayskip=12pt plus 3pt minus 9pt
% \abovedisplayshortskip=0pt plus 3pt
% \belowdisplayskip=12pt plus 3pt minus 9pt
% \belowdisplayshortskip=7pt plus 3pt minus 4pt
% and handle the vertical skip (or space & glue) between paragraph text
% and mathematical equations. The difference between the choice of
% \...skip and \...shortskip depends on whether the preceding/following
% paragraph line is long/short, and accommodates equations that could
% stick out vertically (for example, \sum_{i=1}^k ... in displaystyle).
% If you want these values to only be changed locally (\textit{i.e.}, for a
% specific length of text or equation only), include it within a group:
% \begingroup % or {
% ...
% \begin{equation} % or \[
% ...
% \end{equation} % or \]
% \endgroup % or }
% \vspace{-0.2cm}
% Via establishing the shared semantic space at the video-sentence level and distilling the inferred fine-grained correlation, we expect interpretable attention layers guided by the alignment between video and text in the dataset.
By establishing the shared semantic space at the video-sentence level and distilling the inferred clip-word correlation, we expect correlation-reflective attention layers that also take the fine-grained semantics into account.
% guided by the alignment between video and text in the dataset.
% It essentially guides the attention mechanism to attend to relevant features in the video and text based on the shared semantic understanding.
% \SE{} 
% 1) 글 중간 중간에도 끝에 요약하는거처럼 흐름 잡는 문장들이 들어가면 조금 더 이해가 쉽지 않을까 싶긴 해요
% 예를 들어, guidance G 설명 시, 앞에는 guidance 같은 말이 안나오다가 갑자기 나오는게 사람들 헷갈릴수도 있겠다 싶어요 / 설명 전에 S 등을 통해 video-text align space를 만들었고, 이를 통해 attention map을 더 자세히 guide 하는 방법에 대해 설명한다 이런거?
% 아닌가? 지금이 compact하게 적혀있는거 같기도 하구요...
% 2) G_i 정의할 때, correlation? 인지 cosine-similarity인지? L_distill이 KLD에서 가져온 수식이라면, G의 sum이 1인지 아닌지 이런거에서 사람들이 질문할수도 있을거 같네요 정확히 어떤거였죠? --> correlation, distill할때 bsz, vlen, qlen 이면 qlen축으로 softmax때림.
% 3) 이쪽 내용이 복잡해서 글로도 적지만, 그림에서 각 요소가 어떤 역할을하는지에 대한 설명 같은게 들어가면 이해가 편할거 같아요, 예를 들어, 3.(a) Token Generation에서 \hat{S}, \hat{M} 들이 sentence, video-level moment의 정보를 나타낸다 이런 설명 같은게 그림에 들어가면 어떤지 하는 생각이 드네요

\subsection{Moment-adaptive Saliency Detector}
\label{sec.3.4}
We have discussed ways to calibrate the magnitude of cross-modal interaction concerning the correlation between video and text modalities.
Here, we introduce a novel design of the saliency detector that exploits the calibrated degree of cross-modal interaction in each clip representation.
% \SE{} % intensity gap? clip 별 moment에 기여하는 정도? text query와의 유사성을 사용한다 이런말이죠? + 이 문단이랑 이 아래에서 context에 집중하는거랑 뭔가 핀트가 어긋난거 같아요. 이 intensity gap이 세문단 아래의 clip-wise importance score가 나오기 전에는 언급이 안되는거 같다?
\begin{figure}[t!]
    \centering
    \vspace{-0.2cm}
    \includegraphics[width=1.0\columnwidth]{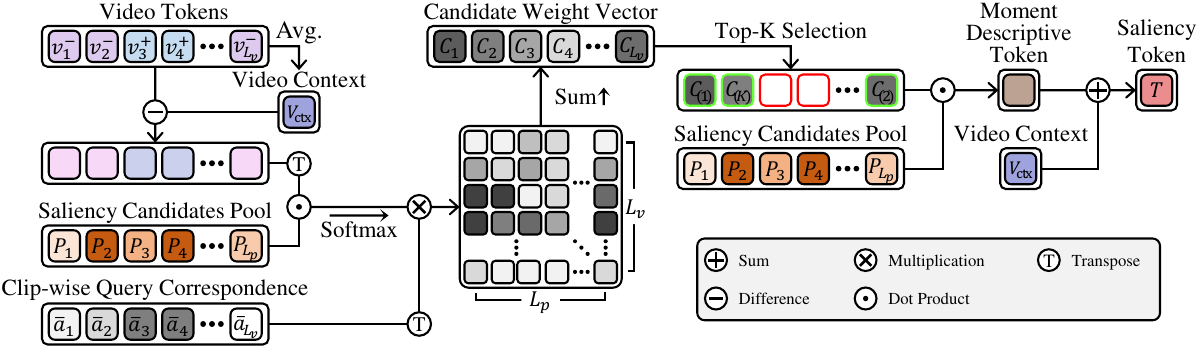}
    \vspace{-0.45cm}
    \caption{
    Saliency token generation. 
    The saliency token is obtained by combining video-averaged context token $V_{\text{ctx}}$ with a moment-descriptive token which is calculated by aggregating top-K moment-descriptive candidates.
    Specifically, we yield a moment-descriptive token by subtracting a context token from clip tokens and then use their correlation to saliency candidates in the pool $P$ as the moment-descriptiveness scores for each candidate.
    Based on these scores after scaling with clip-wise query correspondence $\bar{a}$, we combine Top-K candidates to construct a moment-descriptive token.
    This results in the saliency token that not only maintains contextual similarity with video tokens but also adeptly captures the characteristics of specific moments.
    % This way maintains the saliency token's contextual similarity to video tokens while also being aware of specific moments.
    }
    \vspace{-0.3cm}
    \label{fig:MSD}
\end{figure}

Following \cite{qddetr}, we employ a saliency token to estimate the saliency scores. %by computing the similarity.
This implies that there should be no contextual discrepancy between a saliency token and video features, as the saliency scores are derived from their computed similarity.
% This implies that there should be no contextual discrepancy between a saliency token and video representations since saliency scores are computed with the similarity between them. % based on their similarities.
% This implies that there should be no contextual discrepancy between a saliency token and video representations since saliency scores are computed by measuring the similarity between saliency and video tokens.
In addition, a saliency token also should encapsulate the dynamic characteristics of moments, \textit{e.g.}, actions, within each video for discrimination.

% Following \cite{qddetr}, we employ a saliency token to estimate the saliency scores by computing the similarity.
% This implies that there should be no contextual discrepancy between a saliency token and video representations while a saliency token also should encapsulate the dynamic characteristics of moments, \textit{e.g.}, actions, within each video for discrimination.

Accordingly, we introduce a video context token for minimizing the aforementioned contextual discrepancy, and a moment-descriptive token for encapsulating the diversity of moments.
Then, we generate a saliency token by merging them, as described in Fig.~\ref{fig:MSD}.
% Accordingly, we merge a video context token and a moment-descriptive token to generate the saliency token $T$, as described in Fig.~\ref{fig:MSD}.
In detail, we capture the video context by utilizing the averaged video token so-called a context token $V_{\text{ctx}}$.
Conversely, we generate a moment-descriptive token as a combination of learnable candidates in the candidates pool $P$ following the below steps.
Initially, to effectively incorporate distinct characteristics within each video instance using a finite set of candidate tokens, we start by removing instance-specific context from the video clips.
% To effectively incorporate distinct characteristics of varying moments of every instance using a set number of candidate tokens, we initiate by eliminating instance-wise context from the video clips.
We then perform weighted similarity matching between $L_v$ clip tokens and $L_p$ candidates in the pool to identify moment-representative candidates for specific moments.
Note that the clip-wise query correspondence $\bar{a}$~(defined in Fig.~\ref{fig:textattendance}) are employed as weights in similarity matching to emphasize clips related to desired moments.
The top-K candidates are then aggregated to construct a moment-descriptive token based on their similarity values.
% Subsequently, weighted similarity matching is applied between $L_v$ clip tokens and $L_p$ candidates in the pool. 
Formally, the process is expressed as:
\vspace{-0.3cm}
\begin{equation}
    \label{moment-adatpive saliency detector}
        C_{j} = \sum^{L_v}_{i=1} \bar{a}_i \cdot \frac{\text{exp}\left((v_i - V_{\text{ctx}}) \odot P_j\right)}{\sum_{k=1}^{L_p} \text{exp} \left((v_i - V_{\text{ctx}}) \odot P_k\right)},
\end{equation}
\vspace{-0.35cm}
\begingroup
\setlength{\belowdisplayskip}{-0.1pt}
\begin{equation}
    \label{moment-adatpive saliency detector}
        T = V_{\text{ctx}} + \sum_{j=1}^{L_p} \mathbbm{1}_{C_j \in \{C_{\left(1\right)}, C_{\left(2\right)}, ..., C_{\left(K\right)}\} }P_j \odot C_j,
\end{equation}
\endgroup
where $C_{\left(k\right)}$ denotes the $k$-th biggest value in the candidate weight vector $C$ and $\mathbbm{1}$ is an indicator.

Subsequently, we carry out the projection of the saliency token $T$ through the projector that shares the parameters with the projection layers in the adaptive cross-attention.
In this fashion, we expect the saliency token to be similarly encoded with the video tokens processed via adaptive cross-attention, thereby eliminating the concern of the modality gap between the tokens.

\subsection{Training Objectives}
\label{sec.3.5}
CG-DETR is trained with losses for moment retrieval and highlight detection.
We employ L1, gIoU~\cite{rezatofighi2019generalized}, and cross-entropy for moment retrieval and use margin ranking loss, rank contrastive loss, and entropy loss for highlight detection.
Also, as discussed in Sec.~\ref{aca}, we use the same highlight detection objectives on $\bar{a}$, the clip-wise query correspondence which is derived by adding the attention weights of cross-attention layers.
Abbreviating the moment retrieval, highlight detection, and attention weights $\bar{a}$ learning objectives as $\mathcal{L}_{\text{mr}}$ and $\mathcal{L}_{\text{hl}}$, and $\mathcal{L}_{\text{attn}}$, our overall objective can be formulated as:
\begin{equation}
\label{eq:overall_objective}
\begin{split}
    \mathcal{L}_{\text{obj}} = &\mathcal{L}_{\text{mr}}+\lambda_{\text{hl}} \left( \mathcal{L}_{\text{hl}}+ \mathcal{L}_{\text{attn}} + \mathcal{L}_{\text{bce}} \right) + \lambda_{\text{ortho}} \mathcal{L}_{\text{ortho}} + \lambda_{\text{align}} \mathcal{L}_{\text{align}} + \lambda_{\text{distill}} \mathcal{L}_{\text{distill}}.
\end{split}
\end{equation}
For further details, we refer to the Appendix A.5.

\section{Evaluation}

To validate the generality of our proposed method, we conduct extensive experiments on video grounding datasets.
We use three datasets for moment retrieval: QVHighlights~\cite{momentdetr}, Charades-STA~\cite{gao2017tall}, and TACoS~\cite{tacos}, and three for highlight detection: QVHighlights, TVSum~\cite{song2015tvsum}, and Youtube-hl~\cite{youtubehl}.
For the type of feature extractor, we follow previous works~\cite{umt, momentdetr, qddetr, eatr} for a fair comparison; we utilize Slowfast+CLIP~\cite{CLIP, slowfast}, I3D~\cite{I3D}, and VGG~\cite{simonyan2014very} networks.
% Details for datasets and evaluation metrics are in Appendix A.4.
% We refer to Appendix A.4 for dataset details and evaluation metrics.
% Results for highlight detection benchmarks are in Appendix A.7.
For details on the dataset and evaluation metrics, refer to Appendix A.4. 
The results for highlight detection benchmarks can be found in Appendix A.7.

\subsection{Comparison with the State-of-the-arts}
For joint moment retrieval and highlight detection tasks on QVHighlights, we present comparisons to the state-of-the-art~(SOTA) methods in Tab.~\ref{table_QVHighlight}.
For a fair comparison between methods, numbers are reported with the test and validation splits, respectively~(We leave cells blank if the results are not available).
As observed, our proposed method outperforms previous methods with notable margins, \textit{e.g.}, 8\% boosts in mAP for MR in both splits and 8\% and 12\% increase in R1@0.7.
Our superior performance over previous methods in both tasks substantiates the importance of calibrating the degree of cross-modal interactions. 
Specifically, we believe that the roles of the transformer encoder and decoder become considerably streamlined as they process discretized features based on their relevance to the given query during the cross-modal interaction phase.

% With our superior performance over baselines in both tasks, we state that the importance of calibrating the degree of cross-modal interactions is verified.
% Specifically, we believe that the role of the transformer encoder and decoder is much more simplified because they are to process the discretized features according to the correspondence to the given query during the cross-modal interaction phase.
% By discretizing the query-relevant and irrelevant clips during multimodal interactions, we believe that the role of the transformer encoder and decoder is much more simplified to process the to focus on capturing the moments without 
% may not have to learn the capability to filter out irrelevantly mixed modality features but only exploit these features for precise prediction.
% \SE{} % 여기 살짝 이해가 어려운거 같아요 "may not have to learn the capability to filter out irrelevantly mixed modality features" 이니까 filter out 하는 능력이 없는게 좋다는거죠? irrelevantly mixed modality를 filter out 하면 좋은거 아니예요? 저런 부분에 쓰일 capa가 절약된다 이런뜻인지

\begingroup
\setlength{\tabcolsep}{4.25pt} % Default value: 6pt
\renewcommand{\arraystretch}{1} % Default value: 1
\begin{table*}[t]
\vspace{-0.1cm}
\caption{
% Performance comparison on QVHighlights. Avg. mAP is calculated with IoU thresholds ranging from 0.5 to 0.95 in 0.05 intervals.
Performances on QVHighlights \textit{test} and \textit{val} splits with the features from Slowfast and CLIP. We calculate the average mAP score with IoU thresholds ranging from 0.5 to 0.95 in 0.05 intervals.
}
\label{table_QVHighlight}
\vspace{-0.2cm}
\centering
{\scriptsize 
\begin{tabular}{l|ccccccc|ccccccc}
\hlineB{2.5}
\multicolumn{1}{c|}{Split} & \multicolumn{7}{c|}{test} & \multicolumn{7}{c}{val} \\ \hline
\multicolumn{1}{c|}{\multirow{3}{*}{Method}} &  \multicolumn{5}{c}{MR}     & \multicolumn{2}{c|}{HD} & \multicolumn{5}{c}{MR}     & \multicolumn{2}{c}{HD} \\ \cline{2-15} 
\multicolumn{1}{c|}{} & \multicolumn{2}{c}{R1} & \multicolumn{3}{c}{mAP}  & \multicolumn{2}{c|}{\textgreater{}= Very Good} & \multicolumn{2}{c}{R1} & \multicolumn{3}{c}{mAP}  & \multicolumn{2}{c}{\textgreater{}= Very Good} \\ \cline{2-15} 
\multicolumn{1}{c|}{} & @0.5 & @0.7 & @0.5 & @0.75 & Avg. & mAP & HIT@1 & @0.5 & @0.7 & @0.5 & @0.75 & Avg. & mAP & HIT@1\\ \hlineB{2.5}
B.Thumb~\cite{song2016click}  &  -& -& -& -& -& 14.4 & 20.9 & - & - & - & - & - & - & -\\
DVSE~\cite{liu2015multi}& -& -& -& -& -& 18.8 & 21.8 & - & - & - & - & - & - & -\\
MCN~\cite{anne2017localizing} & 11.4 & 2.7 & 24.9 & 8.2 & 10.7& - & - & - & - & - & - & - & - & -\\
CAL~\cite{escorcia2019temporal} & 25.5 & 11.5& 23.4& 7.7 & 9.9 & - & - & - & - & - & - & - & - & -\\
XML~\cite{lei2020tvr} & 41.8 & 30.4 & 44.6 & 31.7 & 32.1 & 34.5 & 55.3 & - & - & - & - & - & - & -\\
XML+\cite{lei2020tvr} & 46.7 & 33.5 & 47.9 & 34.7 & 34.9 & 35.4 & 55.1 & - & - & - & - & - & - & -\\ \hline
% \multicolumn{9}{c}{DETR decoder} \\ \cline{1-9} 
M-DETR~\cite{momentdetr}& 52.9 & 33.0 & 54.8 & 29.4 & 30.7 & 35.7 & 55.6 & 53.9 & 34.8 & - & - & 32.2 & 35.7 & 55.6 \\
% Moment-DETR w/ PT~\cite{momentdetr} & V& 59.78$_{\pm{0.3}}$& 40.33$_{\pm{0.5}}$& 60.51$_{\pm{0.2}}$& 35.36$_{\pm{0.4}}$& 36.14$_{\pm{0.3}}$& 37.43$_{\pm{0.2}}$ & 60.17$_{\pm{2.7}}$ \\
UMT~\cite{umt} & 56.2 & 41.2 & 53.4 & 37.0 & 36.1 & 38.2 & 60.0 & 60.3 & 44.3 & - & - & 38.6 & 39.9 & 64.2 \\
QD-DETR~\cite{qddetr} & 62.4 & 45.0 & 62.5 & 39.9 & 39.9 & 38.9 & 62.4 & 62.7 & 46.7 & 62.2 & 41.8 & 41.2 & 39.1 & 63.0 \\ 
% QD-DETR&& 62.40$_{\pm_{1.1}}$ & 44.98$_{\pm_{0.8}}$ & 62.52$_{\pm_{0.6}}$ & 39.88$_{\pm_{0.7}}$ & 39.86$_{\pm_{0.6}}$ & 38.94$_{\pm_{0.4}}$& 62.40$_{\pm_{1.4}}$\\ 
% QD-DETR w/ PT & V & 63.18$_{\pm_{1.0}}$ & 45.19$_{\pm_{0.7}}$ & 63.37$_{\pm_{0.7}}$ & 40.35$_{\pm_{0.8}}$ & 39.96$_{\pm_{0.4}}$ & 38.52$_{\pm_{0.1}}$& 61.91$_{\pm_{0.5}}$\\ 
% UMT w/ PT~\cite{umt}& V+A & 60.83& 43.26& 57.33& 39.12& 38.08& 39.12 & 62.39 \\ \hline
% QD-DETR & V+A & 63.06$_{\pm_{1.0}}$ & 45.10$_{\pm_{0.7}}$ & 63.04$_{\pm_{0.9}}$ & 40.10$_{\pm_{1.0}}$ & 40.19$_{\pm_{0.6}}$ & 39.04$_{\pm_{0.3}}$& 62.87$_{\pm_{0.6}}$\\ \hline
UniVTG~\cite{univtg} & 58.9 & 40.9 & 57.6 & 35.6 & 35.5 & 38.2 & 61.0 & 59.7 & - & - & - & 36.1 & 38.8 & 61.8 \\ 
EaTR~\cite{eatr} & - & - & - & - & - & - & - & 61.4 & 45.8 & 61.9 & 41.9 & 41.7 & 37.2 & 58.7 \\
% TR-DETR~\cite{tr-detr} & 64.7 & 49.0 & 64.0 & 43.7 & 42.6 & 39.9 & 63.4 & - & - & - & - & - & - & - \\
\hline
% VT-Former (Ours) & && 64.33 & 46.82 & 55.55 & 38.78 & 37.25 & \textbf{40.92} & 66.15 & 66.13 & 49.1 & 55.57 & 39.61 & 38.01 & \textbf{41.23} & \textbf{66.90} \\
\rowcolor{gray!30}
\textbf{CG-DETR} & \textbf{65.4} & \textbf{48.4} & \textbf{64.5} & \textbf{42.8} & \textbf{42.9} & \textbf{40.3} & \textbf{66.2}& \textbf{67.4} & \textbf{52.1} & \textbf{65.6} & \textbf{45.7} & \textbf{44.9} & \textbf{40.8} & \textbf{66.7} \\ 
\hlineB{2.5}
% \multicolumn{9}{c}{Large scale pretraining} \\ \cline{1-10} 
% UnLoc-B~\cite{unloc} & & \checkmark & - & - & - & - & - & - & - & 64.50 & 48.80 & - & - & - & - & -\\
% UnLoc-L~\cite{unloc} & & \checkmark & - & - & - & - & - & - & - & 66.10 & 46.70 & - & - & - & - & - \\
% UniVTG~\cite{univtg} & & \checkmark & 65.43 & 50.06 & 64.06 & 45.02 & 43.63 & 40.54 & 66.28 & 68.39 & - & - & - & 45.99 & 41.25 & 67.42 \\ \hline
% VT-Former (Ours) & & \checkmark & - &&&&&&& 69.48 & 51.55 & 58.49 & 42.22 & 41.67 & 41.4 & 65.81 \\
% QD-DETRv2 (Ours) &\checkmark & \checkmark & 68.48 & 53.11 & 69.4 & 49.12 & 47.97 & 40.71 & 66.6 & 68.65 & 54.39 & 69.47 & 51.06 & 49.29 & 40.55 & 66.65 \\ 
% \hlineB{2.5}
\end{tabular}
\vspace{-0.3cm}
}
\end{table*}
\endgroup

\begin{table*}[!t]
    \centering
    \scriptsize
    \begin{minipage}[t!]{0.53\linewidth}%\centering
    \setlength{\tabcolsep}{3.pt} % Default value: 6pt
    \renewcommand{\arraystretch}{1.} % Default value: 1
        \caption{Moment retrieval results. Slowfast and CLIP are used for the backbone.
        %Video features are extracted using Slowfast and CLIP.
        }
    \label{table_mr}
     \centering
    {\scriptsize 
    \vspace{-0.25cm}
    \begin{tabular}{l|cccc|cccc}
    \hlineB{2.5}
    \multicolumn{1}{c|}{\multirow{2}{*}{Method}} & \multicolumn{4}{c|}{TACoS} & \multicolumn{4}{c}{Charades-STA}  \\ \cline{2-9} 
    \multicolumn{1}{c|}{}& R0.3 & R0.5 & R0.7 & mIoU & R0.3  & R0.5 & R0.7 & mIoU \\ \hline
    2D-TAN~\cite{2dtan} & 40.0  & 28.0  & 12.9  & 27.2 & 58.8 & 46.0  & 27.5 & 41.3  \\
    VSLNet~\cite{zhang2020span} & 35.5  & 23.5  & 13.2  & 25.0 & 60.3 & 42.7  & 24.1 & 41.6  \\ 
    M-DETR~\cite{momentdetr} & 38.0 & 24.7  & 12.0  & 25.5 & 65.8 & 52.1 & 30.6  & 45.5 \\
    QD-DETR~\cite{qddetr}  & -   & -   & -   & - & -   & 57.3 & 32.6 & - \\
    LLaViLo~\cite{ma2023llavilo} & - & - & - & - & - & 55.7 & 33.4 & - \\
    UniVTG~\cite{univtg} & 51.4 & 35.0  & 17.4  & 33.6 & \textbf{70.8} & 58.0  & 35.7 & 50.1  \\ \hline
    % TR-DETR~\cite{tr-detr} & -   & -   & -   & - & -   & 57.6 & 33.5 & - \\ \hline
    % VT-Former~(Ours) && 49.77 & 36.14 & 19.57 & 33.73 & 69.73 & \textbf{58.63} & 35.16 & 49.72 & 7.23 & \textbf{4.13} & \textbf{1.68} & \textbf{5.06} \\ 
    \rowcolor{gray!30}
    \textbf{CG-DETR} & \textbf{54.4} & \textbf{39.5} & \textbf{23.4} & \textbf{37.4} & 70.4 & \textbf{58.4} & \textbf{36.3} & \textbf{50.1}  \\  
    % \multicolumn{14}{c}{Large Pretraining}         \\ \hline 
    % UniVTG w/ PT  &  & 11.74  & 7.54   & 3.25   & 7.88 & 72.63   & 60.19  & 38.55 & 52.17 & 56.11  & 43.44  & 24.27  & 38.63 \\ 
    % VT-Former~(Ours)& &  - &  &  &  & - &  & & & - &  &  &  \\ 
    % QD-DETRv2~(Ours) &\checkmark& 8.6 & 4.49 & 2.14 & 6.31 & 70.65 & 59.70 & \textbf{40.94} & \textbf{51.67} & 50.81 & \textbf{39.45} & \textbf{24.66} & \textbf{36.29} \\   \hline
    \hlineB{2.5}
    \end{tabular}
    }
    \end{minipage}\hfill%
    \begin{minipage}[t!]{0.46\linewidth}
    \setlength{\tabcolsep}{11.5pt} % Default value: 6pt
    \renewcommand{\arraystretch}{0.92} % Default value: 1
        \caption{Results on Charades-STA with VGG backbone. 
    $\dagger$ indicates usage of audio modality.}
    \label{table_vgg}
    \vspace{-0.25cm}
    	\centering
    {\scriptsize
    \begin{tabular}{l|cc}
    \hlineB{2.5}
    % R1
    Method &  R0.5 & R0.7 \\ \hlineB{2.5}
    SAP~\cite{SAP}    &  27.4  & 13.4 \\
    % TripNet& VGG  & 36.61  & 14.50 \\
    SM-RL~\cite{SM-RL}  &  24.4  & 11.2 \\
    2D-TAN~\cite{2dtan} &  40.9  & 22.9 \\
    FVMR~\cite{gao2021fast}   &  42.4  & 24.1 \\
    APGN~\cite{APGN}   &   44.2  & 25.6 \\
    SSRN~\cite{SSRN}   &   46.7  & 28.0 \\
    UMT$\dagger$~\cite{umt} &  48.3  & 29.3    \\
    QD-DETR~\cite{qddetr} &   52.8  & 31.1 \\ \hline
    % TR-DETR~\cite{tr-detr} & 53.5 & 30.8 \\ 
    % TR-DETR$\dagger$~\cite{tr-detr} & 54.49 & 32.4 \\ \hline 
    \rowcolor{gray!30}
    \textbf{CG-DETR} &  \textbf{55.2} & \textbf{34.2} \\ 
    % Method & feat & Src. & R@0.5 & R@0.7 \\ \hlineB{2.5}
    % SAP~\cite{SAP}    & VGG & V & 27.42  & 13.36 \\
    % % TripNet& VGG  & 36.61  & 14.50 \\
    % SM-RL~\cite{SM-RL}  & VGG & V & 24.36  & 11.17 \\
    % 2D-TAN~\cite{2dtan} & VGG & V & 40.94  & 22.85 \\
    % FVMR~\cite{gao2021fast}   & VGG & V & 42.36  & 24.14 \\
    % APGN~\cite{APGN}   & VGG & V & 44.23  & 25.64 \\
    % SSRN~\cite{SSRN}   & VGG & V & 46.72  & 27.98 \\
    % UMT$\dagger$~\cite{umt} & VGG & V+A & 48.31  & 29.25    \\
    % QD-DETR~\cite{qddetr} & VGG & V & 52.77  & 31.13 \\ \hline 
    % QD-DETRv2 & VGG & V & \textbf{55.22} & \textbf{34.19} \\ 
    \hlineB{2.5}
    \end{tabular}
    % \vspace{-0.1cm}
    }
    \end{minipage}%\hfill
    \vspace{-0.4cm}
\end{table*}
\begingroup
\setlength{\tabcolsep}{6pt} % Default value: 6pt
\renewcommand{\arraystretch}{1} % Default value: 1
\begin{table}[b]
        \vspace{-0.6cm}
        \caption{Component analysis on QVHighlights $\textit{val}$ split.
            ACA, CCL, and MSD denote \textit{adaptive cross-attention}, \textit{clip-word correlation learner}, and \textit{moment-adaptive saliency detector} introduced in Sec.~\ref{aca}, \ref{sec.3.3}, and \ref{sec.3.4}. 
            As ACA can be subdivided into (b) cross-attention, (c) using learnable parameters in cross-attention, (d) projecting parameters into instance-adaptive dummy tokens with dummy encoder~(D.Enc.), and (e) applying losses for dummy tokens to learn query-excluding meaning, we also report ablation results for sub-components of ACA as well.
            % With all components combined in (d), we search the hyperparameters, \textit{i.e.}, number of dummy tokens and number of layers, to train our final model in (e).
            % Ablation studies on adaptive cross attention and saliency detector. 
            % Studies are conducted on QVHighlights \textit{val} split.}
            }
	    \label{table_ablation}
        \vspace{-0.cm}
	\centering
	{\scriptsize
        \begin{tabular}{c|cccc|c|c|ccccccc}
        \hlineB{2.5}
        \multicolumn{1}{c|}{\multirow{2}{*}{}} & \multicolumn{4}{c|}{\multirow{1}{*}{Sec.3.2}} & \multicolumn{1}{c|}{\multirow{1}{*}{Sec.3.3}} & \multicolumn{1}{c|}{\multirow{1}{*}{Sec.3.4}} & \multicolumn{5}{c}{MR} & \multicolumn{2}{c}{HD} \\ \cline{8-14} 
        \multicolumn{1}{c|}{\multirow{2}{*}{}} & \multicolumn{4}{c|}{\multirow{1}{*}{ACA}} & \multicolumn{1}{c|}{\multirow{1}{*}{CCL}} & \multicolumn{1}{c|}{\multirow{1}{*}{MSD}} & \multicolumn{2}{c}{R1}          & \multicolumn{3}{c}{mAP}& \multicolumn{2}{c}{\textgreater{}= V.Good} \\ \cline{2-14} 
        \multicolumn{1}{c|}{} & CA & params & D.Enc. & \multicolumn{1}{c|}{Loss} & \multicolumn{1}{c|}{-} & \multicolumn{1}{c|}{-}
        & @0.5 & @0.7 & @0.5 & @0.75 & Avg. & mAP & HIT@1 \\ \hline
        (a) &&&&&&& 59.9 & 41.9 & 58.7 & 35.9 & 36.0 & 39.1 & 63.2 \\ \hline
        (b) &\checkmark&&&&&& 61.5 & 46.8 & 61.0 & 41.8 & 40.9 & 39.1 & 63.2 \\ 
        (c) &\checkmark& \checkmark&&&&& 63.0 & 48.1 & 62.9 & 43.5 & 43.1 & 39.2 & 62.9 \\  
        (d) &\checkmark& \checkmark &\checkmark &&&& 64.7 & 49.4 & 63.9 & 44.6 & 43.5 & 39.4 & 63.2 \\ 
        (e) &\checkmark& \checkmark &\checkmark &\checkmark &&& 64.9 & 48.7 & 64.5 & 44.7 & 43.5 & 40.2 & 64.9 \\ \hline
        (f) &\checkmark& \checkmark & \checkmark &\checkmark &\checkmark && 66.1 & 49.6 & 65.9 & 45.1 & 44.5 & 40.3 & 65.0 \\ \hline
        (g) &\checkmark& \checkmark & \checkmark &\checkmark &\checkmark &\checkmark& 67.4 & 52.1 & 65.6 & 45.7 & 44.9 & 40.8 & 66.7 \\ \hlineB{2.5}
        \end{tabular}
        \vspace{-0.4cm}
    }
\end{table}
\endgroup

Tab.~\ref{table_mr} and Tab.~\ref{table_vgg} present comparisons on moment retrieval datasets: TACoS and Charades-STA. Interestingly, we observe variations in the performance gap across different datasets compared to previous SOTA methods. 
For instance, while our results are notably superior on QVHighlights, the margin is slightly reduced on TACoS and relatively small on Charades-STA.
We attribute this phenomenon to different dataset characteristics. 
To illustrate, QVHighlights offers long context-rich queries, \textit{e.g.}, \textit{A woman in a blue coat sits and waves to the camera as a man in a lab coat takes a sample.} 
In contrast, Charades-STA features relatively short and simple queries, \textit{e.g.}, \textit{a person sits in a chair}. 
In short, we observe that the performance improvements in our proposed method are more pronounced when diverse textual semantics are provided, \textit{i.e.}, when more unique words with longer queries are given.
For more detailed statistics and discussions, we refer to Appendix A.3.

Finally, CG-DETR does not sacrifice the inference speed despite its superior results on all benchmarks.
Since our clip-word correlation learner is only for training, we highlight the no increase in inference time compared to QD-DETR~\cite{qddetr}.

\subsection{Ablation Studies}
% \SE{} % 여기 위에 단락 간격 원래 이런지?
In Tab.~\ref{table_ablation}, we investigate the benefits of each component.
For baseline (a), we employ the DETR-based architecture~\cite{momentdetr, dabdetr} with a negative pair learning strategy~\cite{qddetr}.
Rows (b) to (g) clearly validate the benefits of each component.
Specifically, from (b) to (e), we dive deep into adaptive cross-attention disclosing the importance of adjusting the degree of cross-modal interaction according to the relevance between video clip and text query: (b) naive cross-attention, (c) using additional \textit{key}s with learnable parameters to allow flexible attention weights~(between 0 and 1) within cross-attention~\cite{ianet}, (d) encoding instance-adaptive dummy tokens with dummy encoder, and (e) learning directly to take attention weights in proportion to the irrelevance between video clips and queries.
(f) further shows that fine-grained relevance consideration beyond sentence level brings additional benefits.
And, finally, with the representations built on the discriminative degree of cross-modal interaction with ACA and CCL, we find that the moment-adaptive saliency detector better captures the highlight clips in (g).
% Also, we briefly give a short review of the impact of the increased capacity of the model in (e) following \cite{univtg, eatr}.
% In addition, we examine the impact of increasing the capacity of the model in (e) following \cite{univtg,eatr}.
For detailed examinations of model capacity w.r.t. the model size, number of dummy tokens, and number of prompt tokens, we refer to extensive ablation studies in Appendix A.6.

\subsection{Analysis}
\begin{figure}[t!]
    \centering
    \vspace{-0.2cm}
    \includegraphics[width=1.0\columnwidth]{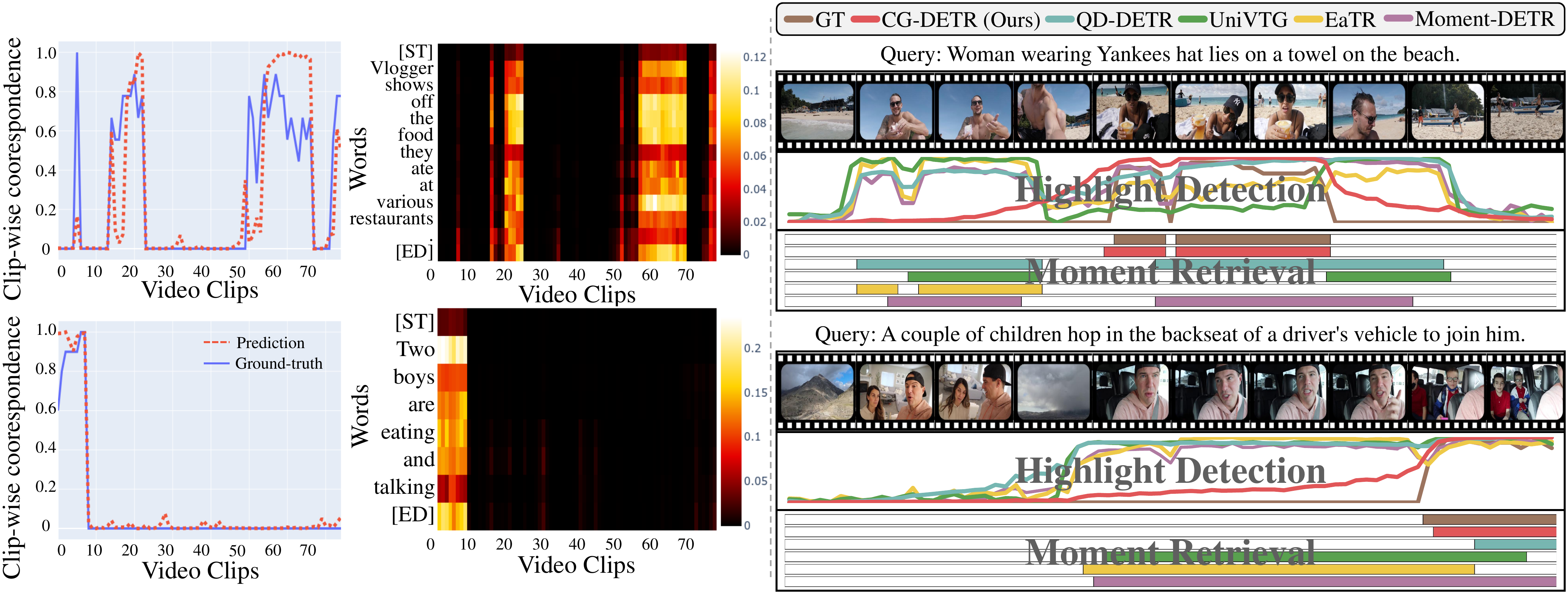}
    \vspace{-0.45cm}
    \caption{
    (Left) Examination of learned correlation between video clips and query in CG-DETR with two video-text pairs~(row).
    Line graphs plot the comparison between the degrees of multimodal interaction and GT saliency scores, and heatmaps show the learned correspondence between each word and video clip. 
    (Right) Visualization of prediction comparisons on QVHighlights.
    }
    \vspace{-0.4cm}
    \label{fig:analysis}
\end{figure}

\textbf{Cross-modal Interaction.}
To examine the cross-modal interaction, we scrutinize how each clip corresponds to the sentence as a whole and each individual word in Fig.~\ref{fig:analysis}~(Left).
On the leftmost side in Fig.~\ref{fig:analysis}, we observe that the degree of interactions is well-calibrated, closely resembling the saliency distribution.
Subsequently, we illustrate the fine-grained interaction degree on its right. 
Although the activations may not be intuitively ranked due to the lack of supervision, we observe that core words such as \textit{food} and \textit{Two} are highly activated.
Also, since the end token $[\text{ED}]$ is treated as the prototype for the text query in CLIP training~\cite{CLIP}, we find that $[\text{ED}]$ is usually highly activated in moment clips.

% Also, as the end token $[\text{ED}]$ is regarded as the prototype for the text query in CLIP training~\cite{CLIP}, we find that $[\text{ED}]$ is usually highly activated on moment clips.

\textbf{Qualitative Results.}
In Fig.~\ref{fig:analysis}~(Right), we plot examples of qualitative results.
To compare against previous SOTA methods, we illustrate the plots with \cite{univtg, momentdetr, qddetr, eatr}.
As more accurate moment predictions against other methods are observed in the bar plot, we believe that the plot shows the significance of scaling the degree of clip-wise cross-modal interaction. 
Furthermore, the line plot showing highlight detection predictions also verifies the effectiveness of exploiting the scaled degree.
For more results, we refer to Appendix A.11.

% \begingroup
% \setlength{\tabcolsep}{2pt} % Default value: 6pt
% \renewcommand{\arraystretch}{1} % Default value: 1
\begin{wraptable}{h}{8.cm}
	% \centering
	{\small 
	\vspace{-0.43cm} 
	\caption{Designs of adaptive cross-attention. Experiments are conducted on QVHighlights \textit{val} split.}
	\label{table_acasink}
    \vspace{-0.25cm}
    \setlength{\tabcolsep}{0.8pt} % Default value: 6pt
    \renewcommand{\arraystretch}{0.95} % Default value: 1
        \begin{tabular}{l|ccccccc}
        \hlineB{2.5}
        \multicolumn{1}{c|}{\multirow{3}{*}{Method}} & \multicolumn{5}{c}{MR} & \multicolumn{2}{c}{HD} \\ \cline{2-8} 
        \multicolumn{1}{c|}{} & \multicolumn{2}{c}{R1} & \multicolumn{3}{c}{mAP} & \multicolumn{2}{c}{\textgreater{}= Very Good} \\ \cline{2-8} 
        \multicolumn{1}{c|}{} & @0.5 & @0.7 & @0.5 & @0.75 & Avg. & mAP & HIT@1 \\ \hlineB{2.5}
        Softmax-One~\cite{softmax} & 63.4 & 47.6 & 62.8 & 43.1 & 42.5 & 39.6 & 63.4 \\ 
        % IANet~\cite{ianet} & 63.0 & 48.1 & 62.9 & 43.5 & 43.1 & 39.2 & 62.9 \\
        Sigmoid & 64.1 & 50.3 & 63.3 & 45.6 & 43.7 & 39.5 & 63.7 \\ 
        Ours & \textbf{67.4} & \textbf{52.1} & \textbf{65.6} & \textbf{45.7} & \textbf{44.9} & \textbf{40.8} & \textbf{66.7}    \\ \hlineB{2.5}
        \end{tabular}
        \vspace{-0.45cm}
        }
\end{wraptable}
% \endgroup
\textbf{Softmax-Off-By-One~(Design of Adaptive Attention).}
Softmax limits the representation of relevance in attention layers by constraining the sum of attention weights to be 1. 
In Tab.~\ref{table_acasink}, we compare our design with alternative designs, \textit{i.e.,} Softmax-One~\cite{softmax, attnsink} and using sigmoid activation.
It is important to note that all other aspects of the design, including the use of $\mathcal{L}_{\text{attn}}$, were kept constant across all experiments.
The results clearly demonstrate the advantages of our ACA design for temporal grounding

% In Table~\ref{table_acasink}, we compare our design with alternative approaches, specifically Softmax-One~\cite{softmax, attnsink} and the use of sigmoid activation. 
% It is important to note that all other aspects of the design, including the use of $\mathcal{L}_{\text{attn}}$, were kept constant across all experiments. The results clearly demonstrate the advantages of our Adaptive Contextual Attention (ACA) design for temporal grounding.

% While the softmax operation hinders the reflection of relevance in cross-attention, other design choices exist to replace na\"ive softmax.
% In Tab.~\ref{table_acasink}, we validate that our design outperforms other designs, \textit{e.g.,} softmax-off-by-one~\cite{softmax, attnsink}, IA-Net~\cite{ianet} and using sigmoid activation.

% Also, by learning per-instance-excluding semantics, dummy tokens learn to encode diverse temporal variations and contexts in the training distribution as a prior for per-instance dummies.
% Similar to detr/slotattention/perceiver, leranable token object context embedding. 
% However, it is the first work to explore the opposite meaning of given...

% \input{figure/figure_5.tex}

% \section{Limitation, Societal Impact, and Conclusion}
\vspace{-0.1cm}
\section{Conclusion }
In this paper, we introduced CG-DETR, calibrating the query-dependency of video representation with respect to the correlation between video clips and text query. 
To enable the calibration of the degree of video-query interaction, we first proposed an adaptive cross-attention equipped with dummy tokens.
By granting dummy tokens a role to take the portion of attention between irrelevant video clips and the text query, we modeled the correlation-guided feature interaction.
% To enable the calibration of the degree of video-query interaction, we first proposed an adaptive cross-attention layer with dummy tokens and guided the feature interaction with correlation.
Then, we devised a clip-word correlation learner to discover fine-grained correlation and further calibrate the interaction at the clip-word level.
% Then, we devised a process to infer the clip-word correlation to further guide the interaction.
Consequently, given the video clip tokens equipped with calibrated query-dependency, we leverage the moment adaptive saliency token to exploit the discrepancy in the degree of interactions.
% Given the calibrated degree of cross-modal interaction with two components, the moment-adaptive saliency token was designed to exploit the discrepancy in the degree of interactions.
With extensive experiments and studies, CG-DETR is verified as a strong tool to associate the desired moments with given language descriptions.
We hope CG-DETR can provide new insights in terms of modality interaction for the video grounding community.

% \vspace{5pt}
% \noindent\textbf{Acknowledgements.} 
{
    \small
    \bibliographystyle{unsrtnat}
    \bibliography{reference}
}
\newpage
\renewcommand{\thesection}{A}   
\renewcommand{\thetable}{A\arabic{table}}   
\renewcommand{\thefigure}{A\arabic{figure}}
\setcounter{section}{0}
\setcounter{table}{0}
\setcounter{figure}{0}
\section{Appendix}
\subsection{Extended Related Work}
\textbf{Vision-text Alignment.}
Driven by the recognition that learning the joint embedding spaces for visual and textual modalities yields effective representations, there has been a surge of interest in vision-text alignment~\cite{blip, blip2, coca, sevila, CLIP, li2017learning, jia2021scaling, GLIP}. 
% Driven by the recognition that the joint learning of shared embedding spaces for visual and textual modalities yields effective representations, there has been a surge of interest in vision-text alignment~\cite{blip, blip2, coca, sevila, CLIP, li2017learning, jia2021scaling, GLIP}. 
% This realization has led to numerous large-scale studies dedicated to exploring the intricacies of vision-text alignment~\cite{CLIP, li2017learning, jia2021scaling, GLIP}. 
Subsequently, outcomes of the studies have been introduced in multimodal downstream tasks, \textit{e.g.}, CLIP has gained prominence for video grounding. 
However, even large pretrained models do not invariably possess a perfect shared embedding space. 
Hence, efforts have been directed toward refining cross-modal interactions in downstream tasks to address such limitations. 
In the context of video grounding, dual contrastive learning~\cite{nan2021interventional} and multimodal reasoning~\cite{zhu2023rethinking} have been proposed to improve text-video alignment. Similarly, QD-DETR~\cite{qddetr} employed negative-pair learning to learn general relationships and fostered active cross-modal interaction. 
Our work strives for a more precise understanding of relations. 
Subsequent to forming an aligned space between the video and sentence, we infer fine-grained clip-word similarity within the space, enabling the manipulation of the behavior of cross-modal interaction for enhanced video temporal grounding.
% practical use와 visual-text modality를 합쳐서 shared embedding space를 학습하면, representation이 좋게 배워진다고 알려져있어, vision-text alignment에 관한 많은 large scale study가 진행되고 있다.~\cite{CLIP, li2017learning, jia2021scaling}
% 이러한 study들은 다양한 multimodal 분야에서 활용되고 있다.
% Video grounding에서는 주로 CLIP encoder가 도입되어 사용되는데, 이러한 large pretrained model들이 항상 완벽한 shared embedding space를 가지고 있지는 않기에, downstream task에서 이를 보완하고자 cross-modal interaction과정을 refine하는 방법들이 있다.
% Video grounding에서도 이러한 방법들이 있는데, ~\cite{nan2021interventional}는 dual contrastive learning을 사용해서 mutual information을 maximize 하는 것을 통해 better align the text and video 하려고 하고, QD-DETR에서는 modality gap 때문에 attention이 modality 간에 되지 않던 문제를 cross-attention 및 general relationship을 배움으로써 해결하려고 하였다.
% 우리는 조금 더 precise한 relation을 배우고자, video-sentence간의 align된 embedding을 학습하고 해당 embedding space에서 fine-grained한 clip-word간의 similarity를 추정하여 cross-attention map의 동작을 manipulate한다.
% ~\cite{QDDETR} 은
% Study on learning image representations with natural language supervision~\cite{CLIP, li2017learning} facilitated the 

% \subsubsection{Detection Transformer}
% Detection transformer~\cite{detr} ...

% 이미지 발전 \cite{groupdetr, msdetr, dacdetr, dabdetr, conditionaldetr, zhu2020deformable}

% video 쪽에는 \cite{yang2022tubedetr, qddetr, momentdetr, he2021end}

\textbf{Hierarchical Interaction.}
Directly aligning vision and text is challenging due to semantic gaps between the modalities.
Several studies have been proposed to leverage hierarchical vision-text interactions to address this issue~\cite{filip, oscar, mvp, pyramidclip}.
OSCAR~\cite{oscar} utilized a pretrained object detector to localize objects and concatenated them with tag information to construct multi-level semantic concepts.
VinVL~\cite{zhang2021vinvl} introduced a pretraining method to enhance the object-attribute detector.
Additionally, MVPTR~\cite{mvp} and X-VLM~\cite{zeng2021multi} utilized multi-level semantic concepts in both visual and textual modalities.
PyramidCLIP~\cite{pyramidclip} achieved remarkable improvements in pretraining by modeling three levels of semantics.
In the video domain, HiVLP~\cite{shao2023hivlp} proposed a hierarchical method to model cross-modal text features while HANet~\cite{wu2021hanet} and UCOFIA~\cite{wang2023unified} utilized hierarchical matching scores for video-text retrieval.
While the primary focus of this body of studies is to organize the hierarchy in learning vision-text alignment, our work primarily aims to reflect the hierarchical vision-text relevance into the attention mechanism using the model's online knowledge.
We believe our approach is complementary to this body of work, and incorporating a hierarchical matching algorithm could further enhance our model for temporal grounding.

% pyramidclip
% PyramidCLIP: Hierarchical Feature Alignment for
% Vision-language Model Pretraining
% image / text 를 global to local / summary to attribute 로 나눠서 학습.

% shao2023hivlp
% wu2021hanet
% Video
% HiVLP: Hierarchical Interactive Video-Language Pre-Training
% hierarchical 하게 cross-modal text feature

% HANet: Hierarchical Alignment Networks for Video-Text Retrieval

% 우리의 방법은 moment-sentence / moment-word를 정렬한다는것에 hierarchical vision-text learning이랑 연관이 깊음.

% Clip wise query correspondence랑 GT saliency score의 similarity의 구간마다 성능과 데이터 수 변화
\begin{figure}[h!]
    \centering
    \includegraphics[width=0.75\textwidth]{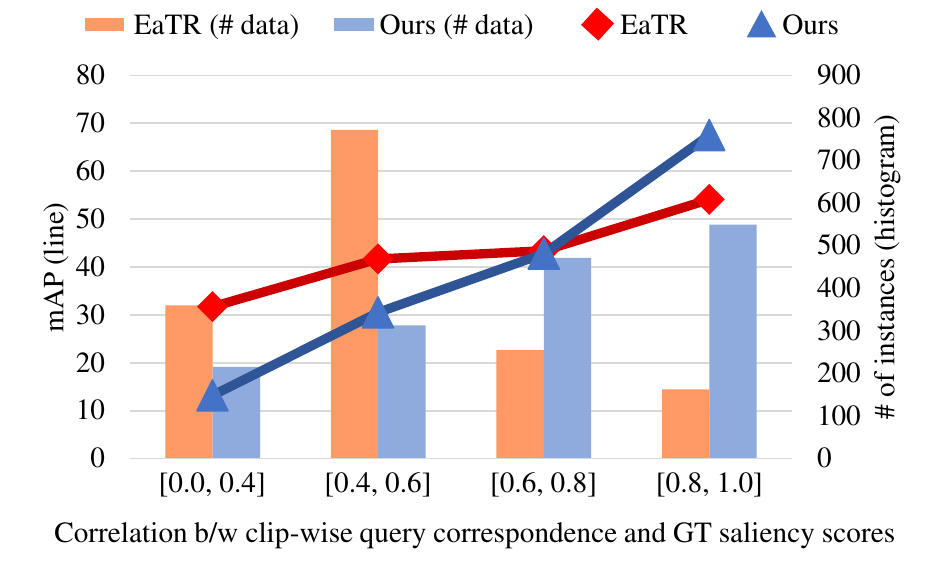}
    \caption{Performance and data count variation across bins of similarity between clip-wise query correspondence~(degree of cross-modal interaction) and GT saliency scores.
    To calculate the similarity between clip-wise-query correspondence and GT saliency scores, we use the cosine similarity. 
    Whereas the histogram displays the number of data points in each interval, the points in the line graph indicate the average score of averaged mAP for moment retrieval in each bin. 
    Our CG-DETR achieves superior performance by appropriately modeling the clip-wise query correspondence~(indicated by the increased numbers in bins with higher similarity) which is a clear indicator of video-text alignment essential for achieving a high mAP score.
    }
    \label{fig:corr_perf_simcorrgt}
\end{figure}
\subsection{Importance of Considering the Degree of Cross-Modal Interaction}
In this section, we illustrate the importance of the degree of cross-modal interaction to support our motivation.
Fig.~\ref{fig:corr_perf_simcorrgt} visualizes the positive correlation between how well clip-wise degrees of cross-modal interaction align with the saliency scores and the performance.
Along the X-axis, we indicate the range of cosine similarity that is calculated between the ground-truth saliency scores and the degree of cross-modal interaction for each video-text paired data.
For each bin in the plot, we calculate the average mAP score for moment retrieval and plot the number of instances that belong to each bin.
As shown, we observe the consistent tendency of performance increase in bins as the degrees of cross-modal interaction resemble the saliency scores which indicates the video-text relevance.
This clearly demonstrates the significance of the degree of cross-modal interaction since it provides clues for how much the text queries correspond to given video clips.
Furthermore, we point CG-DETR's increased number of instances in bins where the similarity between the saliency scores and the degree of cross-modal interaction are higher.
This validates that CG-DETR clearly implements the correlation-reflected interaction by modeling the appropriate correlation during the text-video interaction phase.
Consequently, this leads to the superior performances of CG-DETR.
% In this section, we illustrate the importance of the degree of cross-modal interaction to support our motivation.
% Fig.~\ref{fig:corr_perf_simcorrgt} visualizes the positive correlation between how well clip-wise degrees of cross-modal interaction align with the saliency scores and the performance.
% Along the X-axis, we indicate the range of cosine similarity that is calculated between the ground-truth saliency scores and the degree of cross-modal interaction for each video-text paired data.
% For each bin in the plot, we calculate the average mAP score for moment retrieval and plot the number of instances that belong to each bin.
% As shown with the recent work, EaTR, we observe the tendency of performance increase in bins as the degrees of cross-modal interaction resemble the saliency scores which indicates the video-text relevance.
% By modeling the appropriate correlation during the interaction phase, it is verified with the number of instances in each bin that our model clearly implements the correlation-reflected interaction.
% Furthermore, as the performance drastically increases along with the quality of injected correlation, we believe that our CG-DETR is a more explainable grounding model compared to previous frameworks.

\begin{figure}[t!]
    \centering
    \includegraphics[width=0.65\textwidth]{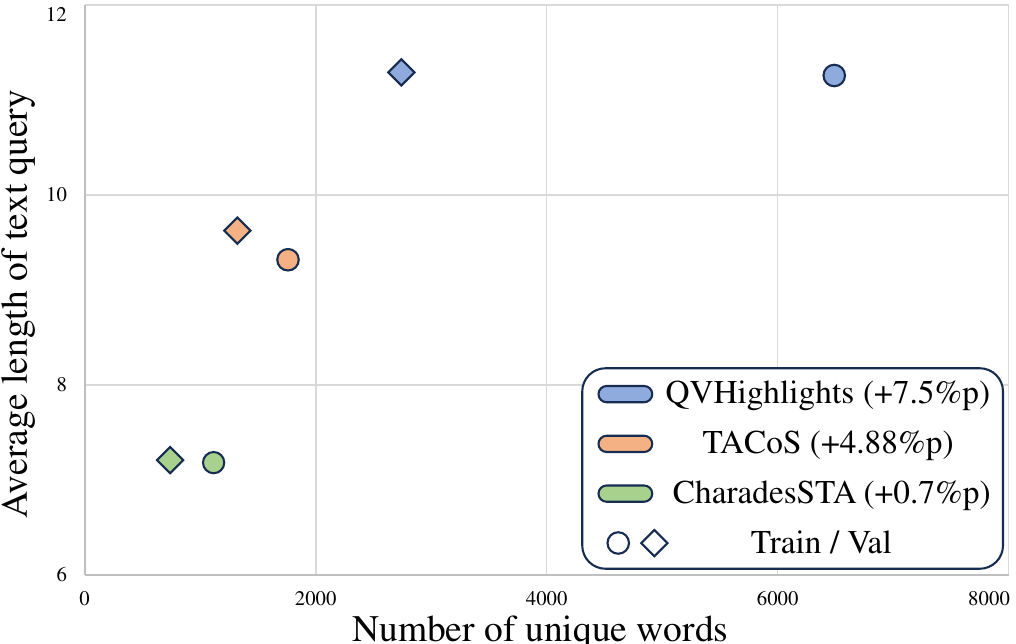}
    \caption{Examined text query complexity of each dataset. X- and Y-axis denote the number of unique words and the average length of text queries in each dataset split, respectively. We state that the query complexity is higher when the number of unique words in each set is bigger and the average query length is longer.}
    \label{fig:fig_dataset}
    % \vspace{-0.2cm}
\end{figure}

\subsection{When does CG-DETR Benefit the Most: Query Complexity of Datasets}
% \begin{figure}[t!]
%     \centering
%     \includegraphics[width=0.65\textwidth]{figure/fig_dataset.pdf}
%     \caption{Examined text query complexity of each dataset. X- and Y-axis denote the number of unique words and the average length of text queries in each dataset split, respectively. We state that the query complexity is higher when the number of unique words in each set is bigger and the average query length is longer.}
%     \label{fig:fig_dataset}
%     % \vspace{-0.2cm}
% \end{figure}
The query complexity of datasets can be defined by the number of semantic words used; we intuitively believe that query complexity is high if diverse semantic words are used and the average query length is longer.
In Fig.~\ref{fig:fig_dataset}, we have plotted the text query complexity of each moment retrieval dataset.
Comparing the query complexity with the moment retrieval performances~(Tab.~1, Tab.~2, and Tab.~3 in the manuscript), we find a strong correlation between the two as indicated in Sec.~4.1 in the manuscript; our moment retrieval results are notable on QVHighlights~(7.5\%p increase in R1@0.7 compared to \cite{univtg}) which possess the highest query complexity and relatively less effective on Charades-STA~(0.7\%p increase in R1@0.7 compared to \cite{univtg}).
% Note that we claim that the dataset has high query complexity when diverse semantic words are used compared to the number of video-text paired instances.
% We observe that the performance improvement gap becomes more pronounced as query complexity increases. 
% Our results are most notable on QVHighlights and relatively less effective on Charades-STA which possess the highest query complexity.
This is because our aim is to discover and apply coarse-to-fine correlations between text and video modalities; when dealing with overly simplistic or excessively short text queries containing fewer semantic components, we anticipate a decrease in benefits. 
% Besides, the query complexity also can be possibly determined by the length of queries in which we observe that the average number of words in the text query of Charades-STA is also much shorter than QVHighlights and TACoS datasets~(the average length of queries in each dataset are 7.2 11.3 and 9.4, respectively).
Consequently, we claim that CG-DETR benefits the most in cases where language descriptions are semantically rich and the model can infer and exploit diverse correlations between queries and videos.

\subsection{Experimental Settings}
\subsubsection{Datasets}
\paragraph{Moment Retrieval.}
\textbf{QVHighlights} is relatively recently publicized dataset by \cite{momentdetr}. Consisting of varying lengths of moments and diverse text queries, it is a challenging and only dataset for joint moment retrieval and highlight detection tasks. Providing 10,310 video-text pairs, it provides a test server on Codalab to ensure fair comparisons.
\textbf{Charades-STA} and \textbf{TACoS} are datasets for moment retrieval.
Charades-STA~\cite{gao2017tall} consists of 9,848 videos regarding indoor activities with an average duration of 30 seconds. 
There exist a few benchmarks depending on the type of feature extractors.
Among them, we follow \cite{qddetr, umt} to test with two popular backbones, \textit{i.e.}, Slowfast+Clip and VGG.
TACoS~\cite{tacos} includes videos mostly about cooking as it is derived from the MPII Cooking Composite Activities dataset~\cite{rohrbach2012script}.
With 127 cooking videos, it has 18,818 moment-query pairs with an average duration of 287 seconds.
Among them, we use 9790 pairs for training and 4436 pairs for testing.

% This finding provides evidence that results may vary depending on the characteristics of the data.

% To provide statistical insights, we have plotted query complexity for each dataset in Fig.~\ref{fig:fig_dataset}.
% Note that we claim that the dataset has high query complexity when diverse semantic words are used compared to the number of video-text paired instances.
% By examining the results alongside query complexity, it becomes evident that there is a strong correlation between the two. 
% We observe that as query complexity increases, the performance improvement gap also becomes more pronounced. 
% This finding provides evidence that results may vary depending on the characteristics of the data.

\paragraph{Highlight Detection.}
\textbf{TVSum} and \textbf{Youtube-HL} are datasets for highlight detection.
TVSUM~\cite{song2015tvsum}, Title-based Video Summarization dataset, composes 50 videos of various genres, \textit{e.g.}, news, documentary, and vlog.
Obtained via crowdsourcing, it has 20 saliency score annotations per video.
We follow the settings in \cite{umt, qddetr}.
YouTube Highlights~\cite{youtubehl} is composed of 433 videos from 6 domains: dog, gymnastics, parkour, skating, skiing, and surfing. We follow \cite{univtg} for the settings, as well as the usage of the domain name as the text query.

\subsubsection{Evaluation Metrics}
To evaluate moment retrieval, we mainly use Recall@1~(R@1) and Mean Average Precision~(mAP) with different IoU thresholds. We also report mean Intersection over Union~(mIoU) for Charades-STA, TACoS, and NLQ datasets.
For highlight detection, we use mAP for QVHighlights, TVSum, and Youtube-hl following~\cite{umt, momentdetr, qddetr, univtg}. HIT@1, a metric to compute the hit ratio for the most highlighted clip, is additionally used for QVHighlights.

% \begingroup
% \setlength{\tabcolsep}{3pt} % Default value: 6pt
% \renewcommand{\arraystretch}{1} % Default value: 1
% \begin{table}[t!]
% \centering
%  % \vspace{-0.1cm}  % 
% 	{\small
% 	\caption{Implementation details. }
% 	\label{table_tvsum_parameter}
% \begin{tabular}{c|cccccccccc}
% \hline
% Domains &  VT        & VU        & GA        & MS        & PK        & PR        & FM        & BK   & BT   & DS \\ \hline
% $M$     & 2 & 1   & 1    & 2  & 2& 2  & 2& 2     & 1    & 2 \\ \hline   
% \end{tabular}}
% \end{table}
% \endgroup

% \begin{table}[t!]
% \centering
%  % \vspace{-0.1cm}  % 
% 	{\small
% 	\caption{Implementation details. }
% 	\label{table_youtube_parameter}
% \begin{tabular}{c|cccccc}
% \hline
% Domains & Dog & Gym. & Park. & Ska. & Ski. & Surf. \\ \hline
% $M$     & 1   & 1    & 2     & 2    & 1    & 2 \\ \hline   
% \end{tabular}}
% \end{table}

\subsection{Training Details}
\label{Sec.trainingdetails}
In this section, we describe the training objectives, hyperparameters, and implementation details.

\subsubsection{Training Objectives}
As elaborated in the manuscript, CG-DETR consists of two sets of objectives: moment retrieval and highlight detection.
To predict the timestamp of the target moments, we utilize $\text{L1}$ and generalized IoU losses~\cite{rezatofighi2019generalized} with cross-entropy~(CE) loss to classify the moment queries between foreground and background.
Let us denote the ground-truth moment and binary classification label as $m = \left(m_c, m_\sigma\right)$ and $y$, and corresponding predictions as $\bar{m} = \left(\bar{m}_c, \bar{m}_\sigma\right)$ and $\bar{y}$, respectively, then the objective is formulated as:
\begin{eqnarray}
    \label{eq_mr_loss}
\mathcal{L}_{\text{mr}}\!=\!\lambda_{\text{L1}}\mathcal{L}_{\text{L1}}(m, \!\bar{m})\!+\!\lambda_{\text{gIoU}}\mathcal{L}_{\text{gIoU}}(m, \!\bar{m})\!+\!\lambda_{\text{CE}}L_{\text{CE}}(y, \!\bar{y}),
\end{eqnarray}
where $\lambda_*$ stands for the coefficients for corresponding losses.

Margin ranking loss, rank contrastive loss, and entropy loss for negative pairs are used for highlight detection.
Among them, margin ranking loss and rank contrastive loss share the objective that ensures the ranking of ground-truth saliency scores is preserved in the predicted scores.
Entropy loss for negative pairs is to suppress the saliency scores of unmatched pairs.
These losses can be formulated as:
\begin{align}
    \label{eq_hl_loss}
    \mathcal{L}_{\text{marg}} = \text{m}&\text{ax}(0, \Delta + S(v_{}^{\text{low}}) - S(v_{}^{\text{high}})) ; \\
    \mathcal{L}_{\text{rctl}} = -\sum_{r=1}^{R}&\text{log}\frac{
    \sum_{v\in V_r^\text{pos}}\text{exp}(S(v)/\tau)}
    {
    \sum_{v\in (V_r^\text{pos} \cup V_r^\text{neg})}\text{exp}(S(v)/\tau)
    } ;\\
    \mathcal{L}_{\text{neg}}& = - \log(1 - S(v^{\text{neg}})),\\
    \mathcal{L}_{\text{hl}}& = \mathcal{L}_{\text{marg}}+\mathcal{L}_{\text{rctl}}+\mathcal{L}_{\text{neg}},
\end{align}
where $\Delta$, $S(\odot)$, $V^{\text{high}}$, $v^{\text{high}}$ and $v_{}^{\text{low}}$ denote a margin, saliency estimation process, and video tokens from pairs of high and low-rank clips, respectively.
$\tau$ is a temperature scaling parameter and $v^{\text{neg}}$ is video token aggregated with unmatched text query.

Consequently, with all the above objectives and the highlight detection losses with attention weights $\mathcal{L}_{\text{attn}}$ combined, our final objective can be formulated as indicated in the manuscript:
\begin{equation}
\label{eq:overall_objective_sup}
\begin{split}
    \mathcal{L}_{obj}=& \mathcal{L}_{\text{mr}}+\lambda_{\text{hl}} \left( \mathcal{L}_{\text{hl}}+ \mathcal{L}_{\text{attn}} + \mathcal{L}_{\text{bce}} \right) + \\ 
    & \lambda_{\text{ortho}} \mathcal{L}_{\text{ortho}} + \lambda_{\text{align}} \mathcal{L}_{\text{align}} + \lambda_{\text{distill}} \mathcal{L}_{\text{distill}}.
\end{split}
\end{equation}

\begingroup
\setlength{\tabcolsep}{1.25pt} % Default value: 6pt
\renewcommand{\arraystretch}{1} % Default value: 1
\begin{table*}[t!]
\centering
 % \vspace{-0.1cm}  % 
\caption{Implementation details. 
    From top to bottom, we enumerate training details for QVHighlights~(QVH.), Charades~(Cha.), TACoS, TVSum~(TVS.), and Youtube-Highlights~(Y-HL.) datasets. 
    SFC, VGG, I3D in the 'Feat' column indicate the type of backbones~(SFC denotes the usage of Slowfast and CLIP).
    From the left to the right, 'bs' denotes the batch size; 'E' denotes the epoch; 'lr' denotes the learning rate; $L_d$ denotes the number of dummy tokens; $L_p$ denotes the number of candidates in the moment-representative candidate pool; K denotes the number of selected candidates to form moment-representative saliency token; Enc denotes the number of transformer encoder layers; Dec denotes the number of transformer decoder layers; ACA denotes the number of adaptive cross attention layers; D.Enc denotes the number of dummy encoder layers; Mom denotes the number of moment encoder layers; Sen denotes the number of sentence encoder layers.}
	\label{table_hyperparameters}
	{\small
\begin{tabular}{l|c|cccccc|cccccc|ccccccc}
\hline
\multicolumn{1}{c|}{\multirow{2}{*}{Data}} & \multicolumn{1}{c|}{\multirow{2}{*}{Feat}} & \multicolumn{6}{c|}{Hyperparameters} & \multicolumn{6}{c|}{Layer} & \multicolumn{7}{c}{Loss}  \\ \cline{3-21}
 &  & bs & E & lr & $L_d$ & $L_p$ & K & Enc & Dec & ACA & D.Enc & Mom & Sen & $\lambda_\text{hl}$ & $\lambda_\text{L1}$ & $\lambda_\text{gIoU}$ & $\lambda_\text{CE}$ & $\lambda_\text{ortho}$ & $\lambda_\text{align}$ & $\lambda_\text{distill}$\\ \hline
QVH. & SFC & 32 & 200   & 1e$^{-4}$ &  45 &10 & 1 & 3 & 3 & 2 & 2 & 1 & 1 & 1 & 10 & 1 & 4 & 1 & 1 & 1\\
Cha. & SFC & 32 & 200   & 2e$^{-4}$ &  45 &10 & 2 & 3 & 3 & 2 & 2 & 1 & 1 & 4& 10 & 1 & 4 & 1 & 1 & 1 \\
Cha. & VGG & 16 & 200   & 2e$^{-4}$ &  45 &10 & 2 & 3 & 3 & 2 & 2 & 1 & 1& 4& 10 & 1 & 4 & 1 & 1 & 1  \\
TACoS & SFC & 32 & 200   & 2e$^{-4}$ & 50 & 10 & 2 & 3 & 3 & 2 & 2 & 1 & 1& 4& 10 & 1 & 4 & 1 & 1 & 1  \\
% NLQ & SF+C    & 32 & 200   & 2e$^{-4}$ & x      & 4      & 1     &10  & 3                   & 3   & 3   & 2   & 2   & 1     & 1    \\
TVS. & I3D & 4 & 1000  & 1e$^{-4}$ & 3 &10 & T.\ref{table_tvsum_parameter} & 3 & 3 & 2 & 2 & 1 & 1 & 4& 10 & 1 & 4 & 1 & 1 & 1 \\
Y-HL. & SFC& 4 & 1000 & 2e$^{-4}$ & 1 &10 & T.\ref{table_youtube_parameter} & 3 & 3 & 2 & 2 & 1 & 1 & 4& 10 & 1 & 4 & 1 & 1 & 1 \\ \hline
\end{tabular}}
% \vspace{-0.1cm}
    % \vspace{-0.1cm}
\end{table*}
\endgroup

\begingroup
\setlength{\tabcolsep}{6pt} % Default value: 6pt
\renewcommand{\arraystretch}{1} % Default value: 1
\begin{table*}[t!]
\centering
    \caption{Number of selected candidates to form moment-representative saliency token for TVSum. }
	\label{table_tvsum_parameter}
{\small
    \begin{tabular}{c|cccccccccc}
    \hline
    Domains &  VT & VU & GA & MS        & PK        & PR        & FM        & BK   & BT   & DS \\ \hline
    K     & 2 & 1   & 1    & 2  & 2& 2  & 2& 2     & 1    & 2 \\ \hline   
    \end{tabular}}
\end{table*}
\endgroup

\begingroup
\setlength{\tabcolsep}{1.5pt} % Default value: 6pt
\renewcommand{\arraystretch}{1} % Default value: 1
\begin{table*}[t!]
        \caption{Number of selected candidates to form moment-representative saliency token for Youtube-hl.}
	\label{table_youtube_parameter}
\centering
{\small
        \begin{tabular}{c|cccccc}
        \hline
        Domains & Dog & Gym. & Park. & Ska. & Ski. & Surf. \\ \hline
        K     & 1   & 1    & 2     & 2    & 1    & 2 \\ \hline   
        \end{tabular}}
\end{table*}
\endgroup

\subsubsection{Hyperparamters}
Parameters for each benchmark are enumerated in Tab.~\ref{table_hyperparameters}.
Following \cite{momentdetr, umt, qddetr} on QVHighlights, Charades-STA, and TVSum, we utilize slowfast~\cite{slowfast}, CLIP~\cite{CLIP}, VGG~\cite{VGG}, and I3D~\cite{I3D} backbone features.
For TACoS and Youtube-HL datasets, we follow \cite{univtg} to use slowfast and CLIP.
Note that we differ the hyperparameter K per domain for highlight detection datasets as different domains are treated as different splits~\cite{univtg}. Details for these datasets are enumerated in Tab.~\ref{table_tvsum_parameter} and Tab.~\ref{table_youtube_parameter}. 
For other details, we use the Adam optimizer with a weight decay of 1e-4 for all experiments.
The hidden dimension of the transformer architecture is set to 256, and $\tau$ for rank contrastive loss is set to 0.5 for all experiments.

\subsubsection{Additional implementation details}
Following \cite{qddetr, univtg}, we also employ a negative-pair learning strategy.
However, to alleviate the risk of forming false negative pairs during training, we do not use the video clips from the same video as a negative pair.
Moreover, as we regard negative pairs to contain any query-corresponding moments, we do not calculate $\mathcal{L}_{\text{distill}}$ with the negative pairs.
Lastly, we also adopt modality embedding following \cite{univtg}.

\begingroup
\setlength{\tabcolsep}{5.1pt} % Default value: 6pt
\renewcommand{\arraystretch}{1} % Default value: 1
\begin{table*}[t!]
\centering
% \vspace{-0.1cm}
{\small
    \caption{Layer ablation. In inference time, only the layers for encoder, decoder, adaptive cross-attention, and dummy encoder are used.}
    \label{table_layer}
    \begin{tabular}{cccccc|ccccccc}
    \hline
    \multicolumn{1}{c}{\multirow{3}{*}{Enc.}} & \multicolumn{1}{c}{\multirow{3}{*}{Dec.}} & \multicolumn{1}{c}{\multirow{3}{*}{ACA}} & \multicolumn{1}{c}{\multirow{3}{*}{D.Enc.}} &\multicolumn{1}{c}{\multirow{3}{*}{Mom}} &\multicolumn{1}{c|}{\multirow{3}{*}{Sent.}} &\multicolumn{5}{c}{MR} & \multicolumn{2}{c}{HD}\\ \cline{7-13}
    & & & & & & \multicolumn{2}{c}{R1} & \multicolumn{3}{c}{mAP} & \multicolumn{2}{c}{\textgreater{}= Very Good} \\ \cline{7-13}
    & & & & & & @0.5 & @0.7 & @0.5 & @0.75 & Avg. & mAP & HIT@1 \\ \hline
    2& 2& 2& 2 & 1 & 1 & 65.29 & 50.32 & 64.21 & 43.47 & 42.79 & 40.23 & 65.87 \\
    2& 2& 2& 2 & 2 & 2 & 64.13 & 48.06 & 64.07 & 42.81 & 42.33 & 39.91 & 64.84 \\
    3& 3& 2& 2 & 1 & 1 & \textbf{67.35} & \textbf{52.06} & \textbf{65.57} & \textbf{45.73} & \textbf{44.93} & \textbf{40.79} & \textbf{66.71} \\
    3& 3& 2& 2 & 2 & 2 & 66.39 & 50.90 & 65.40 & 46.17 & 44.92 & 40.31 & \textbf{66.71} \\
    3& 3& 3& 2 & 2 & 2 & 65.87 & 50.45 & 65.00 & 44.61 & 44.14 & 40.66 & 66.13 \\
    3& 3& 3& 3 & 3 & 3 & 65.29 & 50.90 & 64.85 & 44.27 & 43.69 & 40.62 & 65.87 \\ \hline
    \end{tabular}
    % \vspace{-0.1cm}
    % \vspace{-0.1cm}
    }
\end{table*}
\endgroup

\subsection{Ablation Study}
\label{Sec.ablationstudy}

\subsubsection{Layer}
Previous works have demonstrated the effectiveness of the model capacity for temporal grounding~\cite{eatr, univtg}.
Following them, we also provide the performance variation according to the model size in Tab.~\ref{table_layer}.
As others also observed, we find that the increased capacity in the mainstream, \textit{i.e.}, encoder and decoder, yields higher performances.
Yet, we find that the capacity increase in the moment and sentence encoder may result in decreased performance as these encoders learning the alignment between clearly segmented moments and text queries is a much easier task compared to the grounding task that requires clip-level predictions.
% \SE{} % alignment가 더 쉬운 태스크다 <-- 이거는 이견의 여지는 없는지 좀 궁금하긴 하네요
% 
\begin{figure*}[t!]
    \centering
    \vspace{-0.2cm}
    \includegraphics[width=1.\textwidth]{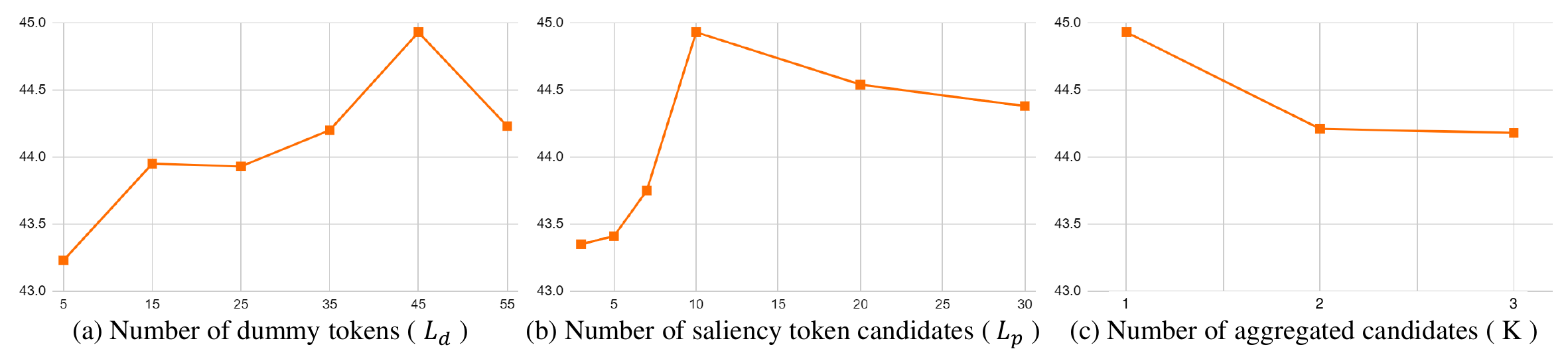}
    \vspace{-0.3cm}
    \caption{Ablation studies on hyperparameters with averaged mAP metric for moment retrieval task.}
    \label{fig:ablation}
    \vspace{-0.2cm}
\end{figure*}

\subsubsection{Dummy tokens}
An ablation study for the number of dummy tokens can be found in Fig.~\ref{fig:ablation}~(a).
Simply put, as the increased number of dummy tokens can express more complicated patterns, we find that there is a tendency for a performance boost with the increased number of dummies.
Yet, we also find that the number of dummies should be determined according to the complexity of patterns in each dataset as too large a number of dummies may disturb the training.
For example, in the case of youtube-hl, we empirically find that setting $L_d$ to 1 performs the best since not only examples in each domain are domain-specific but also do not require fine-grained discrimination with complex patterns.

\subsubsection{Candidate pool for moment-adaptive saliency token}
For the candidate pool, we have two parameters; $L_p$ is the number of token candidates in the pool and $K$ is the number of selected candidates.
We conduct an ablation study for these parameters in Fig.~\ref{fig:ablation}~(b) and (c).
Generally, our model is not very vulnerable to the size of the candidate pool and the number of selected candidates.
Furthermore, we observe that using the maximum combinations with two tokens out of ten candidates can decently express the diverse characteristics of moment clips in the $L_p$ and $K$ columns in Tab.~\ref{table_hyperparameters}.
This validates the idea that matching clips without context helps to efficiently cover varying properties of moment clips with the limited size of the candidate pool.
% \SE{} % 형님 여기 qv에서만 하는거면, N=5 정도도 실험해보는게 어떤가 싶슴다. 전부 높아지면 안좋아지는거처럼 보여서...
% 그리고 마지막 문장이 말하는게, 적당한 수의 N, M (35, 2)이 성능이 제일 좋기 때문에, limited size of pool 이 충분히 moment를 잘 표현한다 이 말인거죠?

\begingroup
\setlength{\tabcolsep}{6.7pt} % Default value: 6pt
\renewcommand{\arraystretch}{1} % Default value: 1
\begin{table*}[h!]
    \caption{Highlight detection results on TVsum. $\dagger$ denotes methods with audio modality. }
	\label{table_TVsum}
    % \vspace{-0.3cm}
	\centering
	{\small
    \begin{tabular}{l|cccccccccc|c}
    \hlineB{2.5}
    \multicolumn{1}{c|}{Method} & VT        & VU        & GA        & MS        & PK        & PR        & FM        & BK   & BT   & DS   & Avg. \\ \hlineB{2.5}
    sLSTM~\cite{zhang2016video}& 41.1      & 46.2      & 46.3      & 47.7      & 44.8      & 46.1      & 45.2      & 40.6 & 47.1 & 45.5 & 45.1 \\
    SG~\cite{mahasseni2017unsupervised} & 42.3      & 47.2      & 47.5      & 48.9      & 45.6      & 47.3      & 46.4      & 41.7 & 48.3 & 46.6 & 46.2 \\
    LIM-S~\cite{xiong2019less}  & 55.9      & 42.9      & 61.2      & 54.0      & 60.3      & 47.5      & 43.2      & 66.3 & 69.1 & 62.6 & 56.3 \\
    Trailer~\cite{wang2020learning} & 61.3      & 54.6      & 65.7      & 60.8      & 59.1      & 70.1      & 58.2      & 64.7 & 65.6 & 68.1 & 62.8 \\
    SL-Module~\cite{hl4} & 86.5      & 68.7      & 74.9      & 86.2     & 79.0      & 63.2      & 58.9      & 72.6 & 78.9 & 64.0 & 73.3 \\ 
    QD-DETR~\cite{qddetr} & \textbf{88.2} & 87.4 & 85.6 & 85.0 & 85.8 & 86.9 & \textbf{76.4} & 91.3 & 89.2 & 73.7 & 85.0 \\
    UniVTG~\cite{univtg} & 83.9 & 85.1 & 89.0 & 80.1 & 84.6 & 81.4 & 70.9 & 91.7 & 73.5 & 69.3 & 81.0 \\ \hline
    MINI-Net~\cite{hong2020mini}$\dagger$ & 80.6 & 68.3 & 78.2      & 81.8      & 78.1      & 65.8      & 57.8      & 75.0 & 80.2 & 65.5 & 73.2 \\
    TCG~\cite{ye2021temporal}$\dagger$ & 85.0      & 71.4      & 81.9      & 78.6      & 80.2      & 75.5      & 71.6      & 77.3 & 78.6 & 68.1 & 76.8 \\
    Joint-VA~\cite{badamdorj2021joint}$\dagger$ & 83.7      & 57.3      & 78.5      & 86.1      & 80.1      & 69.2      & 70.0      & 73.0 & \textbf{97.4} & 67.5 & 76.3 \\
    UMT~\cite{umt}$\dagger$ & 87.5      & 81.5      & 88.2      & 78.8      & 81.4      & 87.0      & 76.0 & 86.9 & 84.4 & \textbf{79.6} & 83.1 \\
    % QD-DETR~\cite{qddetr}  &V+A & 87.6 & \textbf{91.7} & 90.2 & \textbf{88.3} & 84.1 & 88.3 & \textbf{78.7} & 91.2 & 87.8 & 77.7 & 86.6  \\ 
     \hline
     \rowcolor{gray!30}
    \textbf{CG-DETR} & 86.9 & \textbf{88.8} & \textbf{94.8} & \textbf{87.7} & \textbf{86.7}& \textbf{89.6} & 74.8 & \textbf{93.3} & 89.2 & 75.9 & \textbf{86.8} \\ \hlineB{2.5}
    \end{tabular}
    }
    % \vspace{-0.2cm}
\end{table*}
\endgroup

\begingroup
\setlength{\tabcolsep}{9pt} % Default value: 6pt
\renewcommand{\arraystretch}{1} % Default value: 1
\begin{table}[t]
    % \vspace{-0.15cm}
    \caption{Highlight detection results on Youtube-hl. $\dagger$ indicates the use of audio modality. }
	\label{table_youtube}
    % \vspace{-0.0cm}
	\centering
	{\small
    \begin{tabular}{l|cccccc|c}
    \hlineB{2.5}
    \multicolumn{1}{c|}{Method} & Dog & Gym. & Par. & Ska. & Ski. & Sur.  & Avg. \\ \hlineB{2.5}
    RRAE~\cite{yang2015unsupervised} & 49.0 & 35.0 & 50.0 & 25.0 & 22.0 & 49.0 & 38.3 \\
    GIFs~\cite{gygli2016video2gif} &30.8 &33.5& 54.0& 55.4& 32.8& 54.1& 46.4 \\
    LSVM~\cite{youtubehl} &60.0 &41.0& 61.0& 62.0& 36.0& 61.0& 53.6\\
    LIM-S~\cite{xiong2019less} &57.9& 41.7& 67.0& 57.8& 48.6& 65.1& 56.4 \\
    SL-Module~\cite{hl4} &70.8& 53.2 &77.2& 72.5& 66.1& 76.2& 69.3 \\
    QD-DETR~\cite{qddetr} & 72.2 & \textbf{77.4} & 71.0 & 72.7 & 72.8 & 80.6 & 74.4 \\
    UniVTG~\cite{univtg} &71.8 & 76.5 & 73.9 &73.3 &73.2 &82.2 &75.2 \\ 
    \hline
    MINI-Net~\cite{hong2020mini}$\dagger$ & 58.2 &61.7 &70.2 &72.2 &58.7 &65.1 &64.4 \\
    TCG~\cite{ye2021temporal}$\dagger$ &55.4 & 62.7 &70.9 &69.1 &60.1 &59.8 &63.0 \\
    Joint-VA~\cite{badamdorj2021joint}$\dagger$ &64.5 &71.9 &80.8 &62.0 &73.2 &78.3& 71.8 \\
    UMT~\cite{umt}$\dagger$ &65.9 &75.2 &\textbf{81.6}& 71.8& 72.3 &\textbf{82.7}& 74.9 \\
    \hline
    \rowcolor{gray!30}
    \textbf{CG-DETR} &  \textbf{76.3} & 76.1 & 70.0 & \textbf{76.0} & \textbf{75.1} & 81.9 & \textbf{75.9} \\ \hlineB{2.5}
    % \multicolumn{1}{c|}{Method} & Src & Dog & Gym. & Par. & Ska. & Ski. & Sur.  & Avg. \\ \hlineB{2.5}
    % RRAE~\cite{yang2015unsupervised} & V & 49.0 & 35.0 & 50.0 & 25.0 & 22.0 & 49.0 & 38.3 \\
    % GIFs~\cite{gygli2016video2gif} & V &30.8 &33.5& 54.0& 55.4& 32.8& 54.1& 46.4 \\
    % LSVM~\cite{youtubehl} & V &60.0 &41.0& 61.0& 62.0& 36.0& 61.0& 53.6\\
    % LIM-S~\cite{xiong2019less} & V &57.9& 41.7& 67.0& 57.8& 48.6& 65.1& 56.4 \\
    % SL-Module~\cite{xu2021cross} & V &70.8& 53.2 &77.2& 72.5& 66.1& 76.2& 69.3 \\
    % QD-DETR~\cite{qddetr} & V & 72.2 & \textbf{77.4} & 71.0 & 72.7 & 72.8 & 80.6 & 74.4 \\
    % UniVTG~\cite{univtg} &V&71.8 & 76.5 & 73.9 &73.3 &73.2 &82.2 &75.2 \\ 
    % MINI-Net~\cite{hong2020mini} & V+A &58.2 &61.7 &70.2 &72.2 &58.7 &65.1 &64.4 \\
    % TCG~\cite{ye2021temporal} &V+A &55.4 & 62.7 &70.9 &69.1 &60.1 &59.8 &63.0 \\
    % Joint-VA~\cite{badamdorj2021joint} &V+A &64.5 &71.9 &80.8 &62.0 &73.2 &78.3& 71.8 \\
    % UMT~\cite{umt} &V+A &65.9 &75.2 &\textbf{81.6}& 71.8& 72.3 &\textbf{82.7}& 74.9 \\
    % \hline
    % QD-DETRv2 & V & \textbf{76.3} & 76.1 & 70.0 & \textbf{76.0} & \textbf{75.1} & 81.9 & \textbf{75.9} \\ \hlineB{2.5}
    \end{tabular}
    % \vspace{-0.15cm}
    }
\end{table}
\endgroup

\subsection{Results on Highlight Detection Benchmarks}
Results for highlight detection benchmarks are reported in Tab.~\ref{table_TVsum} and Tab.~\ref{table_youtube}.
% Unlike the moment predictions made with decoder outputs, saliency scores for highlight detection are calculated with the encoder outputs.
% Unlike the moment prediction head attached to the decoder, we calculate the saliency scores with the encoder outputs.
Not only does our work achieve SOTA results on average compared to previous methods that utilize only the video modality but also exceeds methods that use additional audio modality.
For drops in a few domains, we point out the unreliability of comparing domain by domain since each domain is only a small subset of data with huge domain gaps.

\begingroup
\setlength{\tabcolsep}{3.5pt} % Default value: 6pt
\renewcommand{\arraystretch}{1} % Default value: 1
\begin{table*}[t!]
\caption{Results with large-scale pretraining on QVHighlights \textit{test} and \textit{val} splits. Pretraining was conducted on Ego4D and VideoCC datasets.
        }
	\label{table_pt}
	\centering
	{\small 
	% \vspace{-0.1cm}  % 
        \begin{tabular}{l|c|ccccccc}
        \hlineB{2.5}

\multicolumn{1}{c|}{\multirow{3}{*}{Method}} & \multicolumn{1}{c|}{\multirow{3}{*}{PT}} &  \multicolumn{5}{c}{MR}     & \multicolumn{2}{c}{HD} \\ \cline{3-9} 
\multicolumn{1}{c|}{} & \multicolumn{1}{c|}{} & \multicolumn{2}{c}{R1} & \multicolumn{3}{c}{mAP}  & \multicolumn{2}{c}{\textgreater{}= Very Good}  \\ \cline{3-9} 
\multicolumn{1}{c|}{} & \multicolumn{1}{c|}{} & @0.5 & @0.7 & @0.5 & @0.75 & Avg. & mAP & HIT@1\\ \hlineB{2.5}
\multicolumn{9}{c}{Validation split}\\ \hlineB{2.5}
UniVGT~\cite{univtg} & \xmark & 59.74 & - & - & - & 36.13 & 38.83 & 61.81 \\ 
CG-DETR (Ours) & \xmark & \textbf{67.35} & \textbf{52.06} & \textbf{65.57} & \textbf{45.73} & \textbf{44.93} & \textbf{40.79} & \textbf{66.71} \\ \hline  \hline
UniVTG~\cite{univtg} & \checkmark & 68.39 & - & - & - & 45.99 & \textbf{41.25} & \textbf{67.42} \\ 
CG-DETR (Ours) & \checkmark & \textbf{68.65} & \textbf{54.39} & \textbf{69.47} & \textbf{51.06} & \textbf{49.29} & 40.55 & 66.65 \\ \hlineB{2.5}
\multicolumn{9}{c}{Test split}\\ \hlineB{2.5}
UniVGT~\cite{univtg} & \xmark &  58.86 & 40.86 & 57.6 & 35.59 & 35.47 & 38.20 & 60.96 \\
CG-DETR (Ours) & \xmark &\textbf{65.43} & \textbf{48.38} & \textbf{64.51} & \textbf{42.77} & \textbf{42.86} & \textbf{40.33} & \textbf{66.21} \\ \hline
% \hlineB{2.5}
% \multicolumn{9}{c}{Large scale pretraining} \\ \cline{1-10} 
% UnLoc-B~\cite{unloc} & & \checkmark & - & - & - & - & - & - & - & 64.50 & 48.80 & - & - & - & - & -\\
% UnLoc-L~\cite{unloc} & & \checkmark & - & - & - & - & - & - & - & 66.10 & 46.70 & - & - & - & - & - \\
% VT-Former (Ours) & & \checkmark & - &&&&&&& 69.48 & 51.55 & 58.49 & 42.22 & 41.67 & 41.4 & 65.81 \\
% VT-DETR (Ours) & (I) & 63.68 & 46.56 & 64.19 & 41.87 & 41.81 & 39.13 & 63.49 & 65.10 & 49.42 & 64.34 & 43.74 & 42.96 & 39.50 & 64.52 \\ 
\hline
UniVTG~\cite{univtg} & \checkmark & 65.43 & 50.06 & 64.06 & 45.02 & 43.63 & 40.54 & 66.28 \\
CG-DETR (Ours) & \checkmark & \textbf{68.48} & \textbf{53.11} & \textbf{69.40} & \textbf{49.12} & \textbf{47.97} & \textbf{40.71} & \textbf{66.60} \\
\hlineB{2.5}
\end{tabular}
\vspace{0.1cm}% 
}
\end{table*}
\endgroup

\subsection{Large-scale Pretraining}
\label{Sec.pretraining}
Recently, as many works have observed the effectiveness of pretraining strategies, a lot of attention is being paid to pretraining strategies.
In this subsection, we scrutinize whether our CG-DETR benefits from the pretraining.
Particularly, we adopt the recent pretraining technique for temporal grounding~\cite{univtg}.
We use Ego4D~\cite{ego4d}, VideoCC~\cite{videocc} datasets with unified annotations~ provided by UniVTG~\footnote{https://github.com/showlab/UniVTG}.
Pretraining was conducted on 8 NVIDIA Tesla A100 GPUs.

Results are reported in Tab.~\ref{table_pt}.
As can be seen, our proposed method gains superior performances after pretraining.
Compared to the very recent baseline, CG-DETR achieves 10\% and 7.2\% performance gains with the mAP metric for moment retrieval.
Furthermore, ours also shows relatively fewer symptoms of overfitting whereas the performances of the baseline differ in \textit{test} and \textit{val} splits.

\begingroup
\setlength{\tabcolsep}{4pt} % Default value: 6pt
\renewcommand{\arraystretch}{1} % Default value: 1
\begin{table*}[t!]
\caption{Performance comparison between ours with convolutional decoder and DETR on QVHighlights \textit{test} and \textit{val} splits.
        }
	\label{table_conv_qv}
	\centering
	{\small
	% \vspace{-0.1cm}  % 
        \begin{tabular}{l|ccccccc}
        \hlineB{2.5}
\multicolumn{1}{c|}{\multirow{3}{*}{Method}} &  \multicolumn{5}{c}{MR}     & \multicolumn{2}{c}{HD} \\ \cline{2-8} 
\multicolumn{1}{c|}{} & \multicolumn{2}{c}{R1} & \multicolumn{3}{c}{mAP}  & \multicolumn{2}{c}{\textgreater{}= Very Good} \\ \cline{2-8} 
\multicolumn{1}{c|}{} & @0.5 & @0.7 & @0.5 & @0.75 & Avg. & mAP & HIT@1 \\ \hlineB{2.5}
\multicolumn{8}{c}{Validation split} \\ \hline
Convolution~(Ours) & 66.13 & 49.1 & 55.57 & 39.61 & 38.01 & \textbf{41.23} & \textbf{66.90} \\ 
DETR~(Ours) & \textbf{67.35} & \textbf{52.06} & \textbf{65.57} & \textbf{45.73} & \textbf{44.93} & 40.79 & 66.71 \\
\hline
\multicolumn{8}{c}{Test split} \\ \hline
Convolution~(Ours) & 64.33 & 46.82 & 55.55 & 38.78 & 37.25 & \textbf{40.92} & 66.15  \\
DETR~(Ours) & \textbf{65.43} & \textbf{48.38} & \textbf{64.51} & \textbf{42.77} & \textbf{42.86} & 40.33 & \textbf{66.21} \\
\hlineB{2.5}
\end{tabular}
% \vspace{-0.1cm}% 
}
\end{table*}
\endgroup

\begingroup
\setlength{\tabcolsep}{6pt} % Default value: 6pt
\renewcommand{\arraystretch}{1} % Default value: 1
\begin{table*}[t!]
    \caption{Performances of ours with convolutional decoder and DETR on NLQ. Video features are extracted using Slowfast and CLIP.}
    \label{table_conv_nlq}
	\centering
	{\small 
    \begin{tabular}{l|cccc}
    \hlineB{2.5}
    \multicolumn{1}{c|}{\multirow{2}{*}{Method}} & \multicolumn{4}{c}{NLQ} \\ \cline{2-5} 
    \multicolumn{1}{c|}{}& R@0.3 & R@0.5 & R@0.7 & mIoU  \\ \hline
    Conv-based & \textbf{7.33} & \textbf{4.16} & \textbf{2.01} & \textbf{5.21} \\ 
    DETR-based & 6.07 & 3.10 & 1.19 & 4.71  \\   \hline
    \hlineB{2.5}
    \end{tabular}
    }
\end{table*}
\endgroup

\subsection{Decoder for Moment Retrieval: Convolution v.s. DETR}

Recent temporal grounding techniques can be categorized into convolution-based and DETR-based.
In this subsection, we briefly study the benefits of two types of architecture.
In short, each type has its strengths in different circumstances.
As can be expected intuitively, the convolution-based model has its advantage when processing the datasets with the property of locality whereas the DETR-based is superior in making predictions for more complex datasets.
To verify the claim, we implement and compare both versions on QVHighlights and NLQ datasets where their variations in moment lengths vary; QVHighlights is challenging due to the large variation in the length of moments varies from pair to pair and moment length in NLQ does not vary much.

Results are in Tab.~\ref{table_conv_qv} and Tab.~\ref{table_conv_nlq}.
While the results with DETR architectures are more powerful in more challenging QVHighlights dataset, the convolution decoder shows its strength when locality exists in moment length.
To provide detailed statistics for the locality, the standard deviation of moment length is 0.237 for QVHighlights whereas it is 0.046 for NLQ~(We normalized every video length to 1 and calculated the standard deviation of portions of the moment).
Results for highlight detection with QVHighlights are comparable because they share the encoder architecture where predictions for highlight detection are yielded.

% \SE{} % 형님 이거도 혹시 다른 데이터셋도 저렇게 moment std와 성능이 일관성이 있었나요? conv vs detr 과 moment의 std

% Since we mainly focus on cross-modal interactions for temporal grounding, our proposed method is equipped with applicability with different decoder architectures.
% Hence, we conduct experiments with both heads and study the strength of each decoder type.
% As can be noticed further in the subsequent performance comparison subsection, simply put, we find that methods with convolution decoder show their strengths in datasets where the lengths of the moments do not vary much, \textit{\textit{e.g.}}, NLQ, whereas the power of DETR can be exploited in more challenging datasets.
% We assume the difference in strengths comes from the locality in moment lengths.
% For the rest of the paper, we name our framework either VT-Former or VT-DETR according to the decoder design.

% As can be easily thought, convolution and DETR have their own benefits. 
% Whereas the convolutional head has its strength in predicting moments in datasets where the length of the moment is similar, \textit{\textit{e.g.}}, NLQ, , and , the power of DETR can be exploited in cases where moments vary from short to long clips, \textit{\textit{e.g.}}, QVHighlights.

% ## QV, NLQ dataset moment variation 보여주면서.
% conv detr 비교.

\subsection{Limitation} 
\label{Sec.limitation}
To fully exploit the strong capabilities of CG-DETR, context-rich text queries and text-clip relevance scores need to be available.
Given that the core advantage of CG-DETR lies in the exploration of coarse-to-fine correlations between video content and text descriptions, its effectiveness may be constrained within datasets of limited complexity.
Despite these challenges, CG-DETR has achieved state-of-the-art results in current benchmarking datasets.
To easily exploit its power and expand to more diverse scenarios, our future research will focus on deducing text-clip relevance scores autonomously, without relying on annotated data.

\subsection{Negative Societal Impact} 
\label{Sec.broaderimpact}
Advances in moment retrieval systems, while intended to facilitate efficient video understanding and analysis, could inadvertently pave the way for erosions of anonymity and individual privacy if such technologies are misused or deployed without proper safeguards. 
Although the core objective of CG-DETR is not to enable unauthorized access to personal or private video data, the capability to accurately localize and retrieve specific video moments raises concerns about potential exploitation for unwarranted surveillance and non-consensual facial recognition.
Therefore, researchers and practitioners must have an ethical responsibility to consider the potential misuse of this work.
\subsection{Visualizations}
\begin{figure*}[t!]
    \centering
    \includegraphics[width=0.84\columnwidth]
    {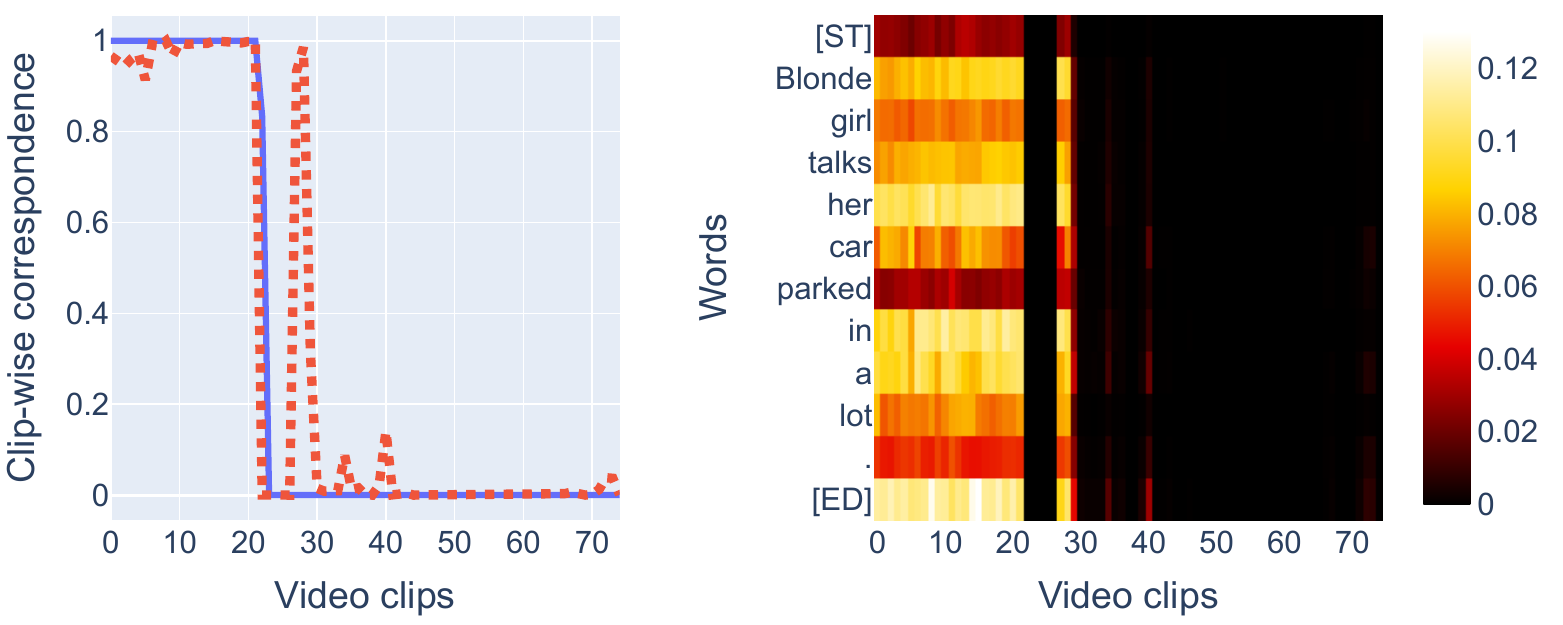}
    \vspace{0.6cm}
    \includegraphics[width=0.84\columnwidth]{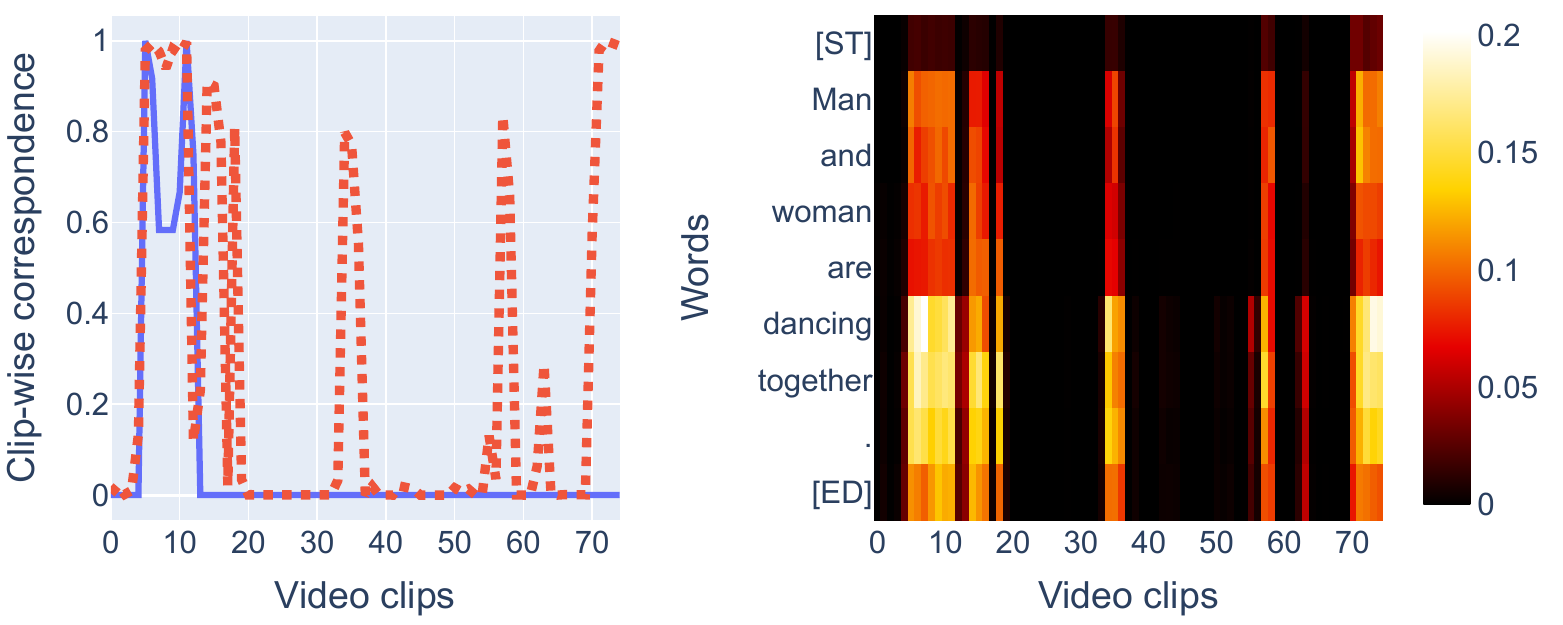}
    \vspace{0.6cm}
    \includegraphics[width=0.84\columnwidth]{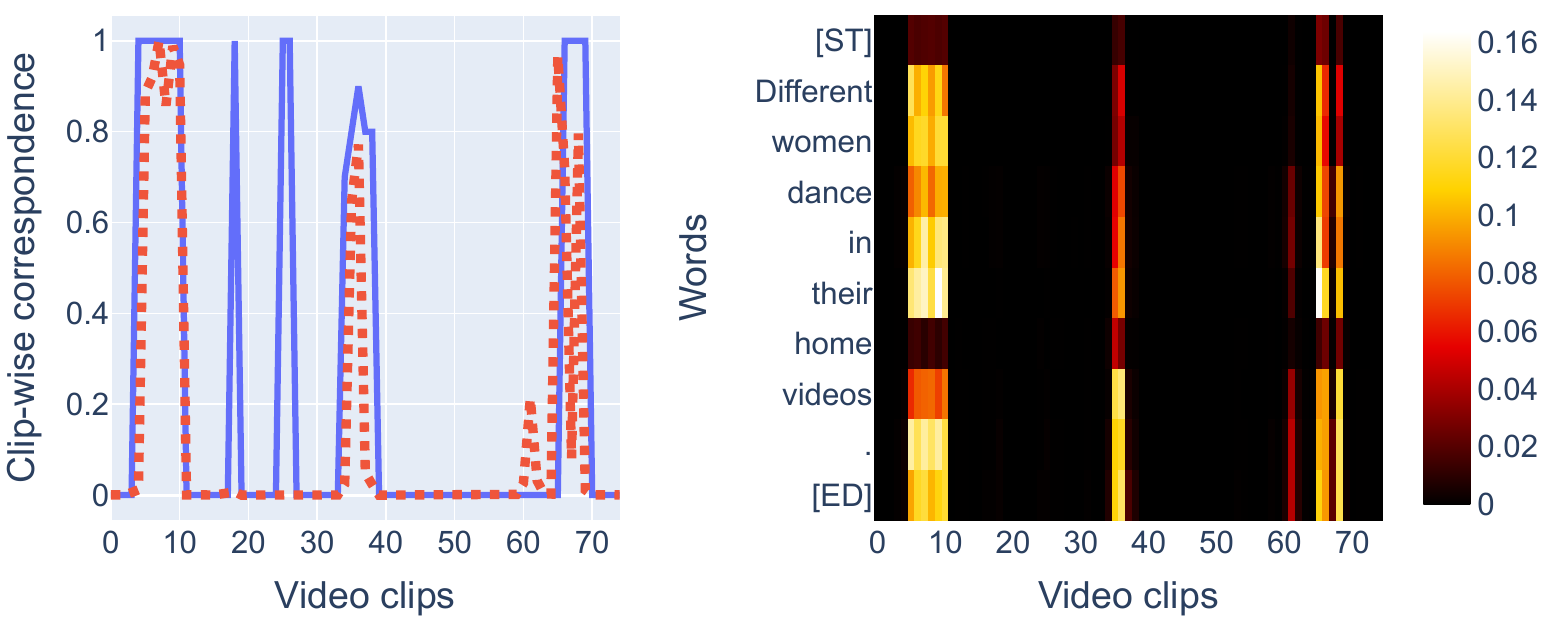}
    \vspace{0.6cm}
    \includegraphics[width=0.84\columnwidth]{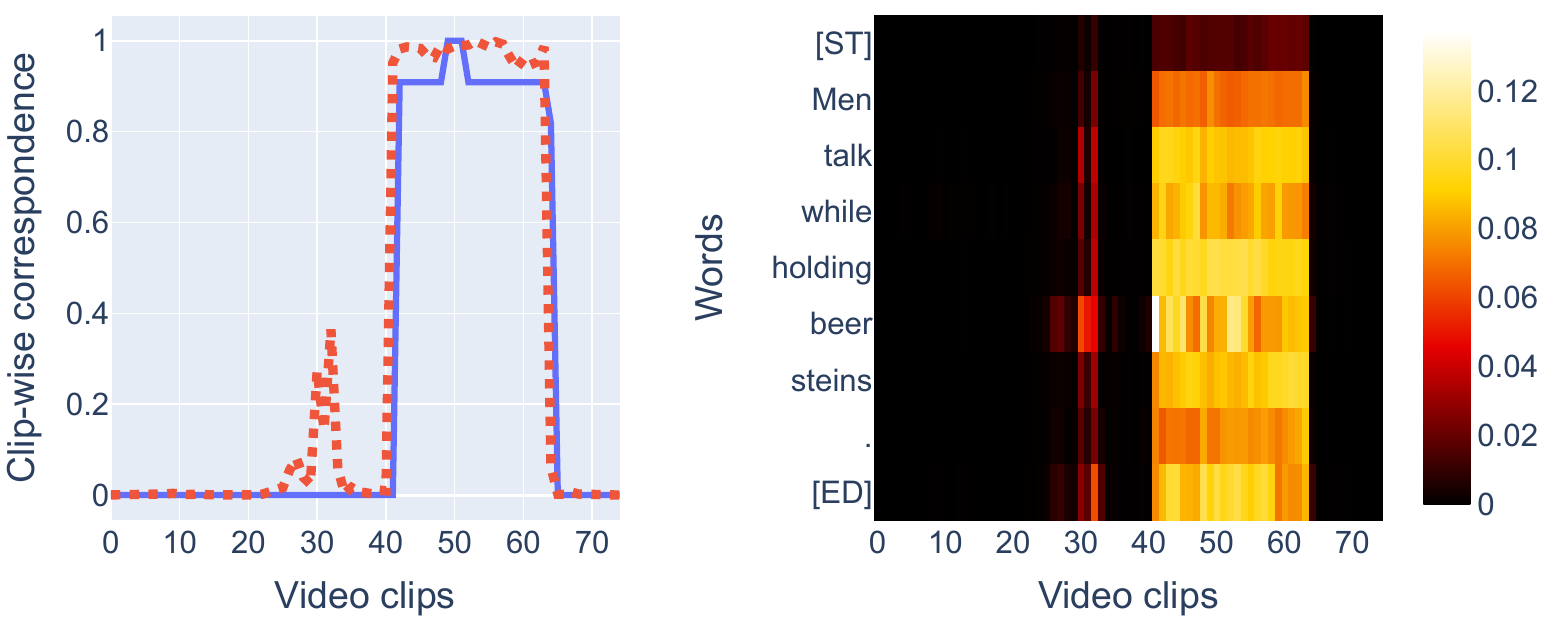}
    \vspace{0.6cm}
    \caption{Visualizations of learned correlation between paired video clips and text queries.}
    \label{fig:word_attn_supp1}
\end{figure*}

\begin{figure*}[t!]
    \centering
    \includegraphics[width=0.84\columnwidth]{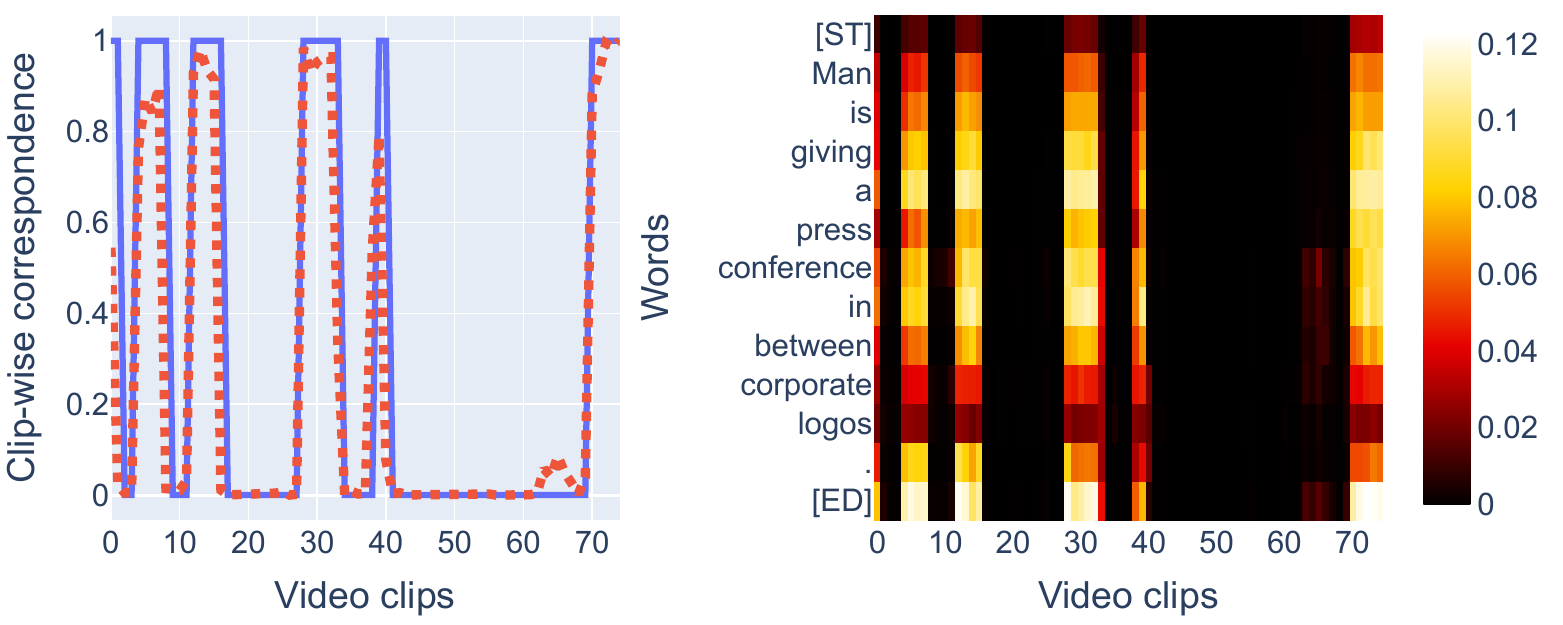}
    \vspace{0.6cm}
    \includegraphics[width=0.84\columnwidth]{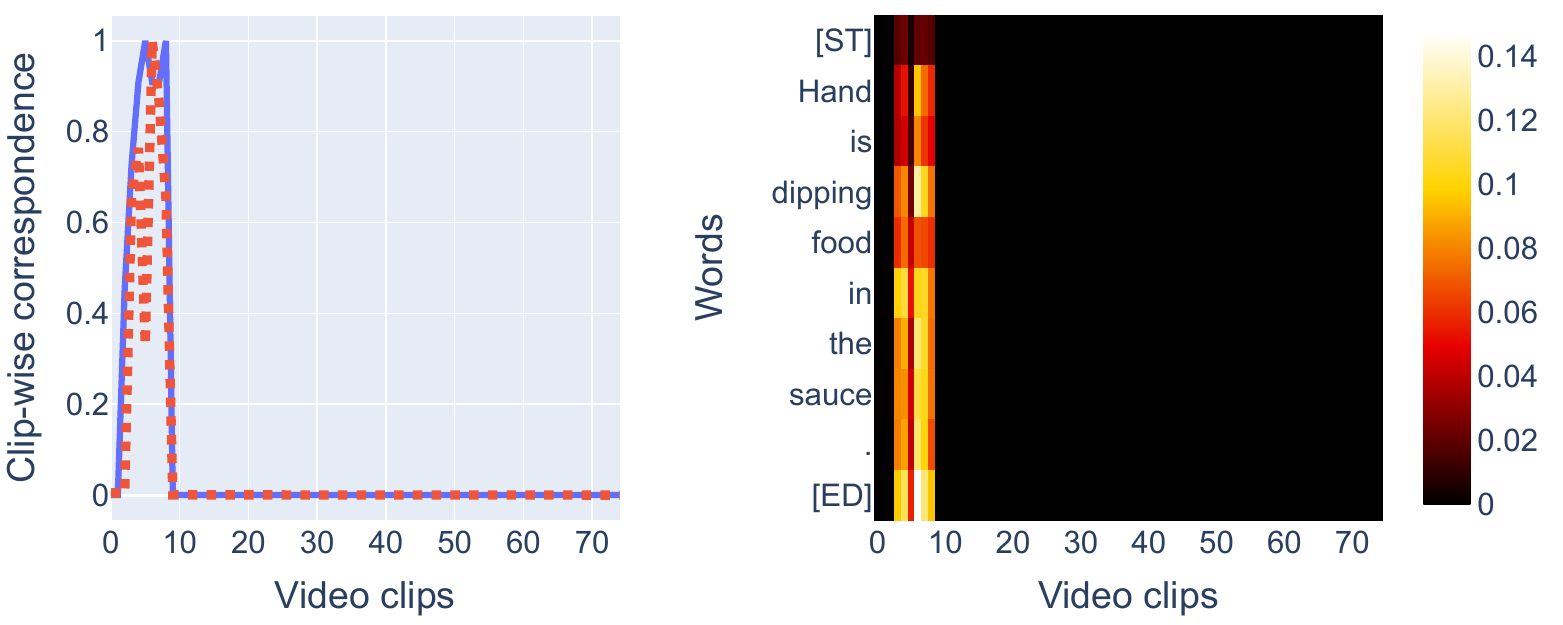}
    \vspace{0.6cm}
    \includegraphics[width=0.84\columnwidth]{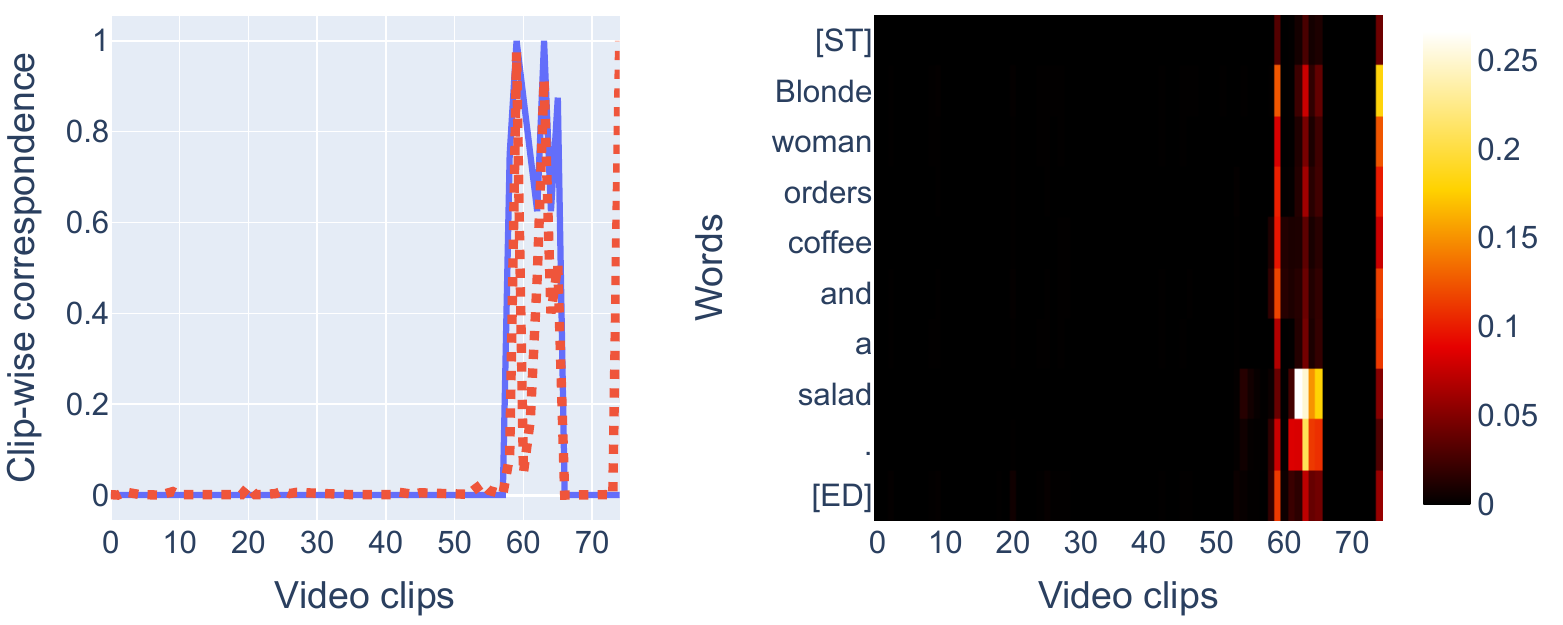}
    \vspace{0.6cm}
    \includegraphics[width=0.84\columnwidth]{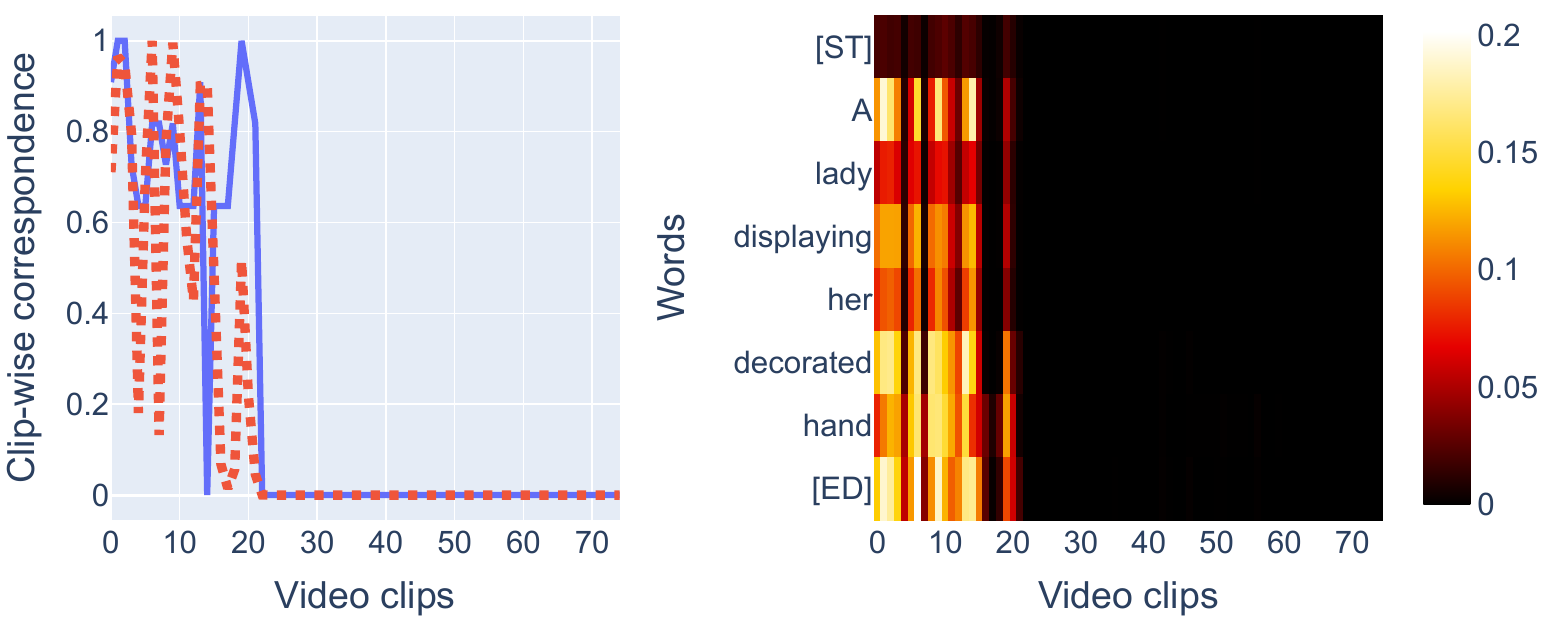}
    \vspace{0.6cm}
    \caption{Visualizations of learned correlation between paired video clips and text queries.}
    \label{fig:word_attn_supp2}
\end{figure*}

\begin{figure*}[t!]
    \centering
    \includegraphics[width=0.84\columnwidth]{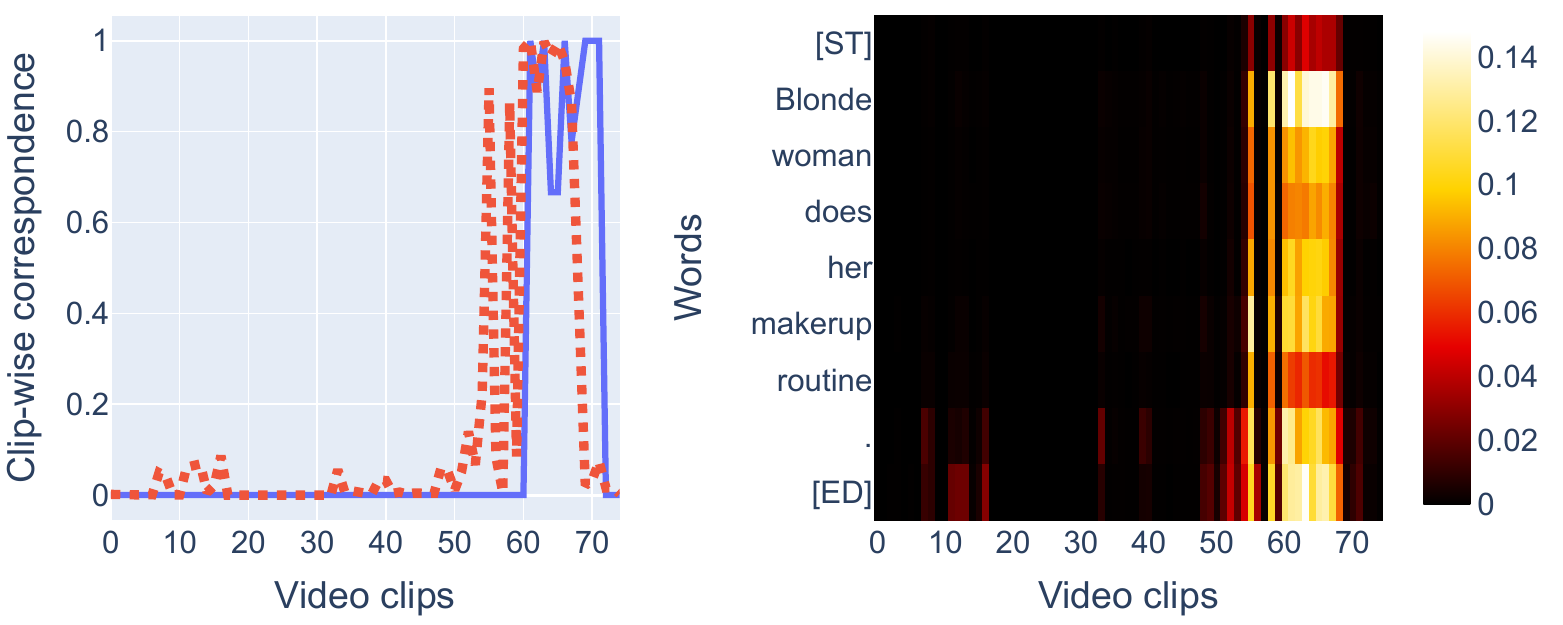}
    \vspace{0.6cm}
    \includegraphics[width=0.84\columnwidth]{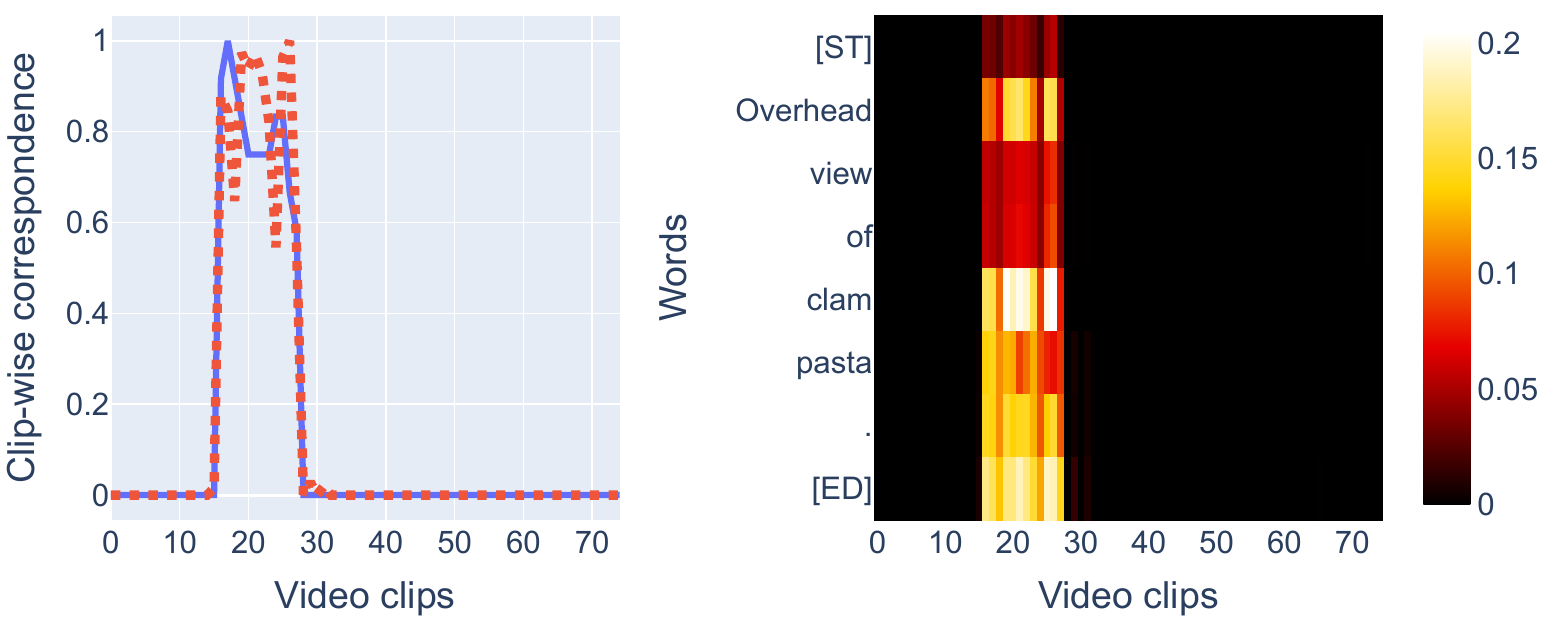}
    \vspace{0.6cm}
    \includegraphics[width=0.84\columnwidth]{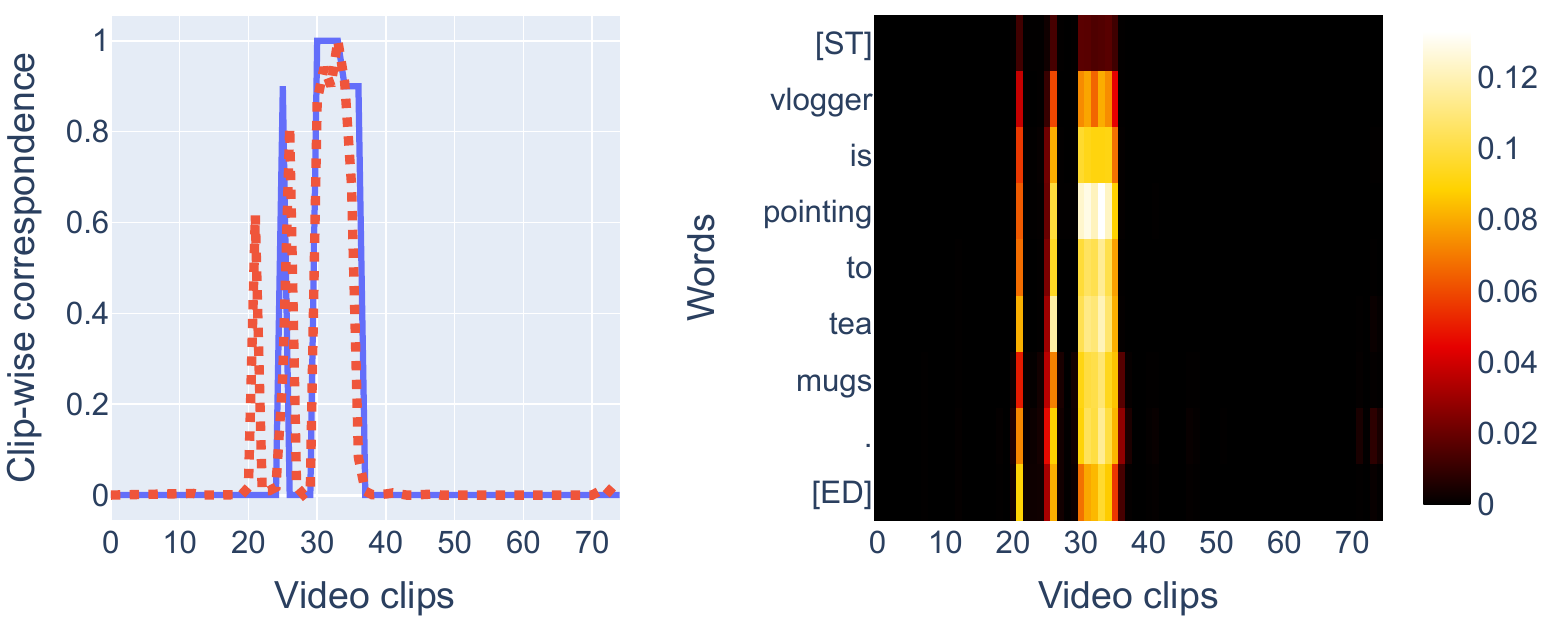}
    \vspace{0.6cm}
    \includegraphics[width=0.84\columnwidth]{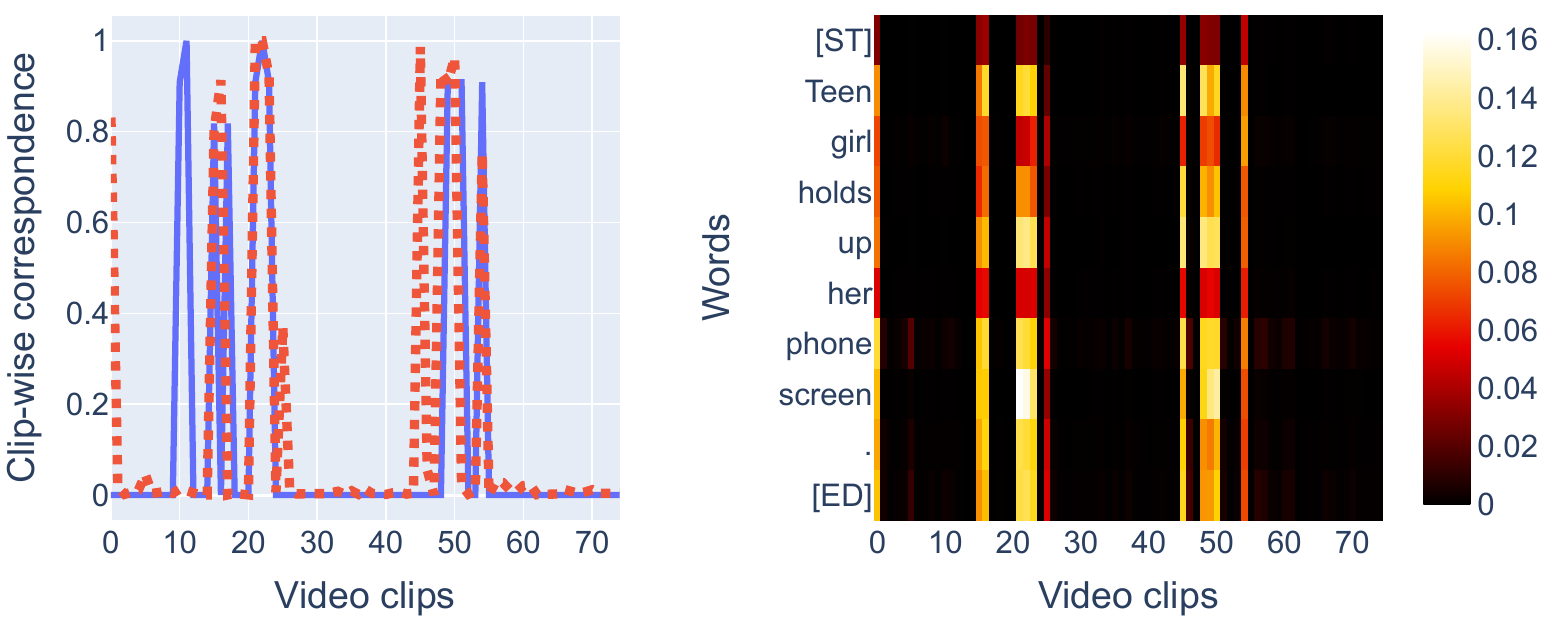}
    \vspace{0.6cm}
    \caption{Visualizations of learned correlation between paired video clips and text queries.}
    \label{fig:word_attn_supp3}
\end{figure*}

\begin{figure*}
    \centering
    \includegraphics[width=1.\textwidth]{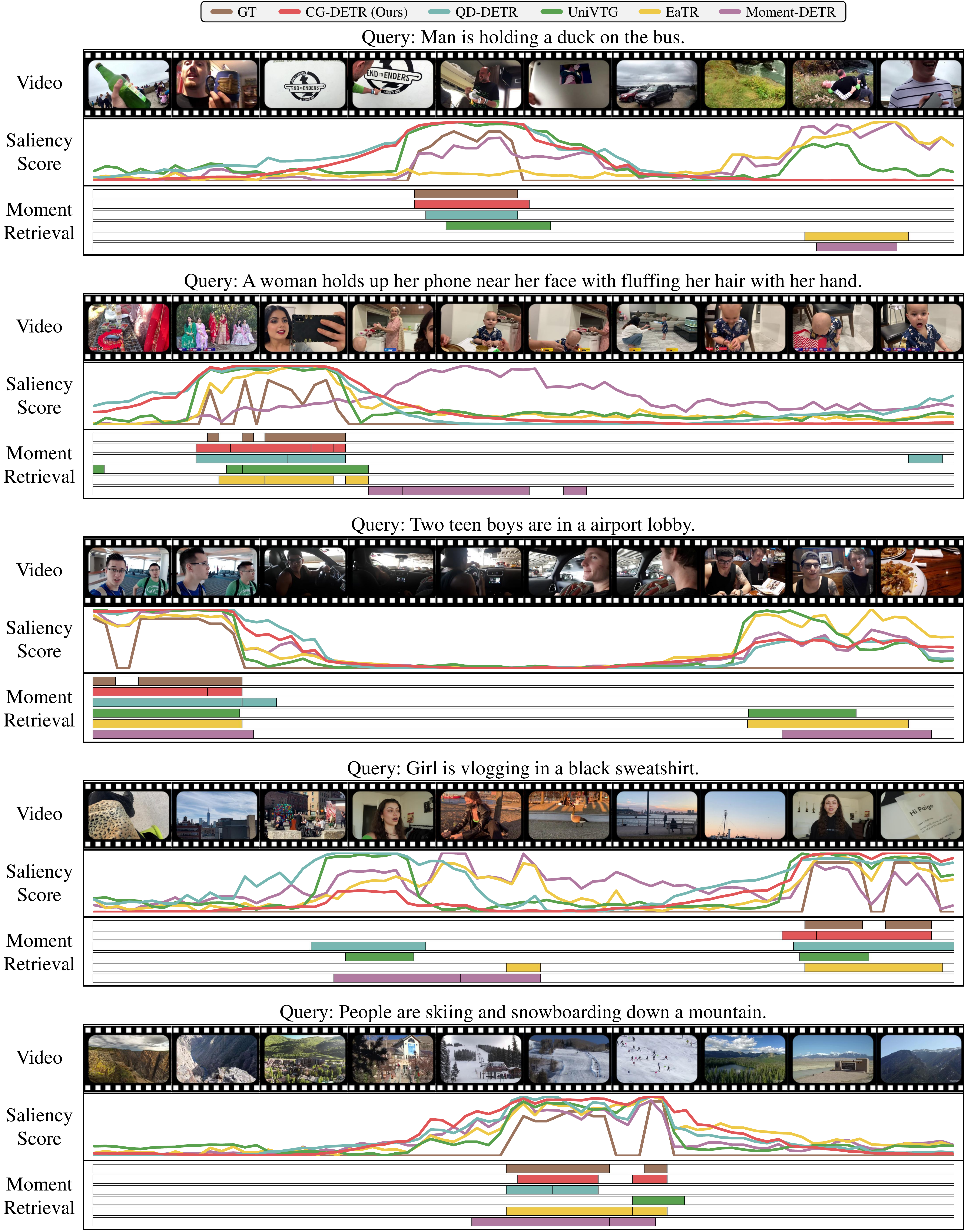}
    \vspace{-0.3cm}
    \caption{Qualitative results.}
    \label{appendix_qualitative1}
\end{figure*}

\begin{figure*}
    \centering
    \includegraphics[width=1.\textwidth]{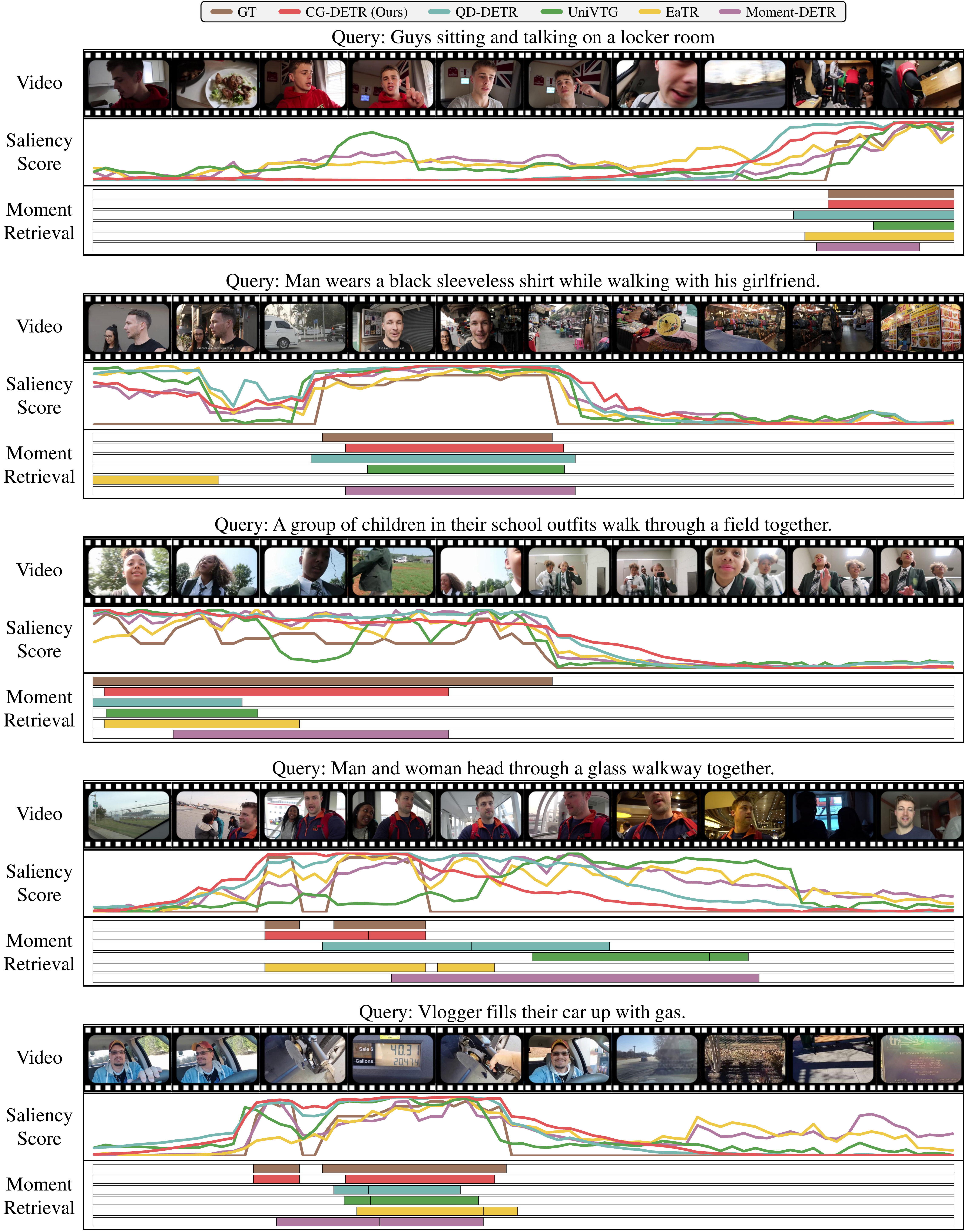}
    \vspace{-0.3cm}
    \caption{Qualitative results.}
    \label{appendix_qualitative2}
\end{figure*}

\begin{figure*}
    \centering
    \includegraphics[width=1.\textwidth]{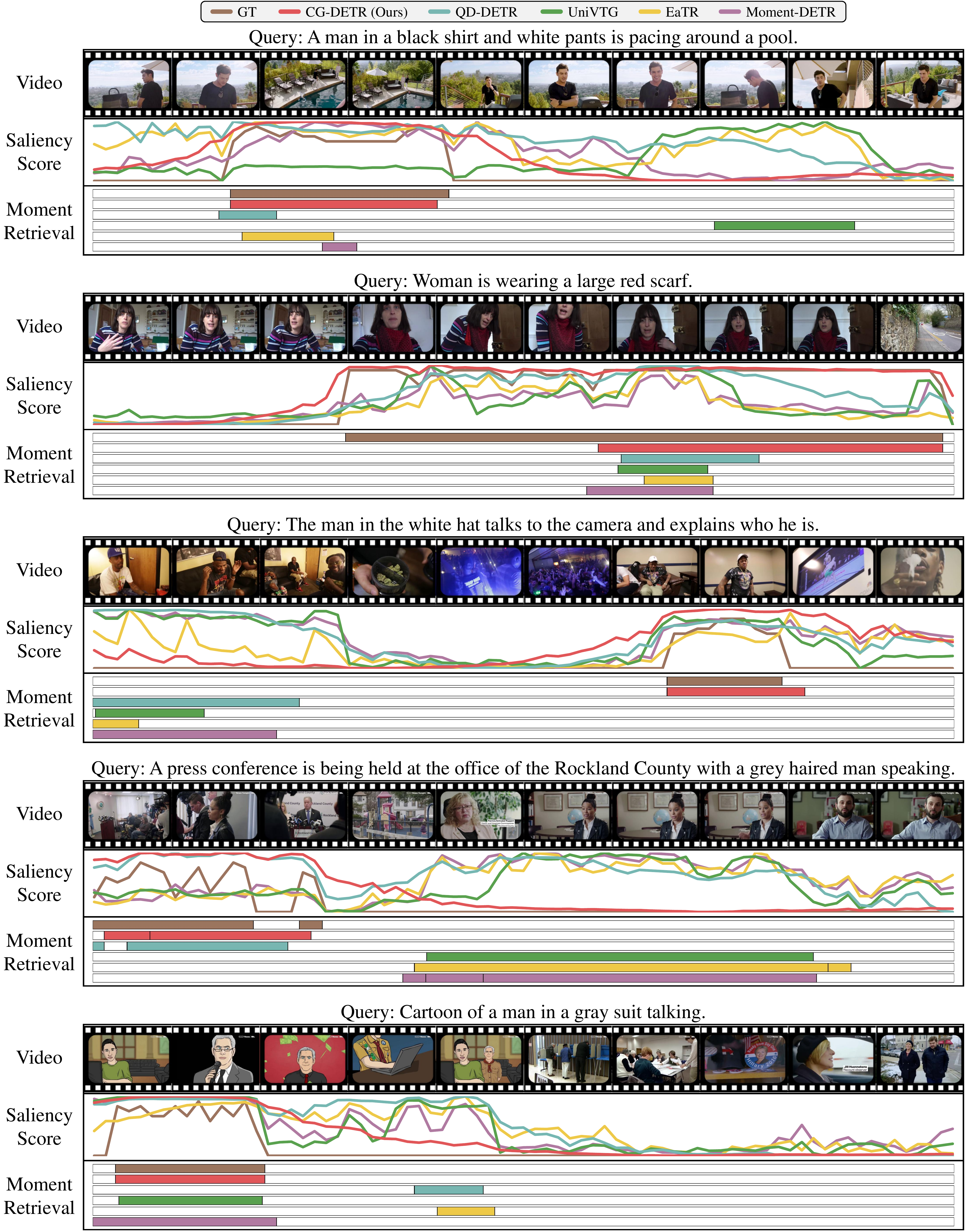}
    \vspace{-0.3cm}
    \caption{Qualitative results.}
    \label{appendix_qualitative3}
\end{figure*}

% \vspace{-1cm}
In Fig.~\ref{fig:word_attn_supp1}, Fig.~\ref{fig:word_attn_supp2}, and Fig.~\ref{fig:word_attn_supp3}, we show additional paired plots of the learned correlation between video clips and text queries at sentence and word levels.
Fig.~\ref{appendix_qualitative1}, Fig.~\ref{appendix_qualitative2}, and Fig.~\ref{appendix_qualitative3} visualize additional qualitative comparisons between methods on the QVHighlights dataset.

% \vspace{-0.3cm}
% \vspace{-0.3cm}

\newpage
\newpage
\newpage

\end{document}